\documentclass[preprint,12pt]{elsarticle}
\usepackage{fullpage}
\usepackage{media9}
\addmediapath{./Movies/}
\usepackage{multimedia}
\usepackage{subcaption}
\usepackage[english]{babel}
\usepackage[utf8]{inputenc}
\usepackage{amsmath}
\usepackage{amssymb}
\usepackage{float}
\usepackage{bm}
\usepackage{multicol}
\usepackage{dirtytalk}
\usepackage{graphicx}
\usepackage{caption}
\usepackage[margin=1in]{geometry}
\usepackage{graphicx}
\usepackage[colorinlistoftodos]{todonotes}
\usepackage{algorithm}
\usepackage{algpseudocode}

\DeclareMathOperator*{\argmin}{argmin}
\usepackage{setspace}

\newcommand{\bk}{\mathbf{k}}
\newcommand{\bx}{\mathbf{x}}

\newcommand{\bY}{\mathbf{Y}}
\newcommand{\bI}{\mathbf{I}}
\newcommand{\bK}{\mathbf{K}}

\newcommand{\btheta}{\boldsymbol{\theta}}

\begin{document}
\begin{frontmatter}
\title{Reinforcement Learning based Sequential Batch-sampling for Bayesian Optimal Experimental Design} 
\author[ual]{Yonatan Ashenafi}
\author[ge]{Piyush Pandita} 
\author[ge]{Sayan Ghosh}

\address[ual]{University of Alberta, Edmonton, T6G 2R3 AB, Canada}
\address[ge]{General Electric Research, Niskayuna, New York, 12309, United States}

\begin{abstract}
Engineering problems that are modeled using sophisticated mathematical methods or are characterized by expensive-to-conduct tests or experiments, are encumbered with limited budget or finite computational resources.
Moreover, practical scenarios in the industry, impose restrictions, based on logistics and preference, on the manner in which the experiments can be conducted.
For example, material supply may enable only a handful of experiments in a \emph{single-shot} or in the case of computational models one may face significant wait-time based on shared computational resources.
In such scenarios, one usually resorts to performing experiments in a manner that allows for maximizing one's state-of-knowledge while satisfying the above mentioned practical constraints.
Sequential design of experiments (SDOE) is a popular suite of methods, that has yielded promising results in recent years across different engineering and practical problems.
A common strategy, that leverages Bayesian formalism is the Bayesian SDOE, which usually works best in the one-step-ahead or myopic scenario of selecting a single experiment at each step of a sequence of experiments.
In this work, we aim to extend the SDOE strategy, to query the experiment or computer code at a batch of inputs.
To this end, we leverage deep reinforcement learning (RL) based policy gradient methods, to propose batches of queries that are selected taking into account entire budget in hand.
The algorithm retains the sequential nature, inherent in the SDOE, while incorporating elements of reward based on task from the domain of deep RL.
A unique capability of the proposed methodology is its ability to be applied to multiple tasks, for example optimization of a function, once its trained.
We demonstrate the performance of the proposed algorithm on a synthetic problem, and a challenging high-dimensional engineering problem.
\end{abstract}
\end{frontmatter}

\section{Introduction}
Design of experiments (DOE)~\cite{chernoff, bartroff} is an essential aspect of most, if not all, engineering tasks. 
Tasks ranging from choosing geometric parameters for aircraft rotors to pharmacological drug design~\cite{liu2021drugex} require careful experimentation. 
An optimally designed experiment is expected to be reliable, to give useful information, and be cost effective. 
These objectives are often taxing and at odds with each other. 
Optimal experimental design~\cite{atkinson2007optimum,box1992sequential} effectively addresses these challenges. 
One important challenge where this is demonstrated, is in the optimization of black-box functions that are expensive to evaluate (e.g. to run simulations or lab experiments on the propeller efficiency of rotors with given geometric configurations)~\cite{Jones1998}. 

Optimizing black-box simulators, that are expensive to evaluate, is an open research problem. 
Bulk of the literature, involves the Bayesian optimization suite of methods~\cite{Jones1998}, that have shown promise in being able to identify the optimal inputs to a simulator with single or multiple objectives~\cite{emmerich2018tutorial}.
In recent years, a plethora of research has focused on Bayesian design of experiments~\cite{beck2018fast,long2013fast,long2015fast,long2015laplace,long2014projection}, both one-shot designs~\cite{tsilifis2017efficient,ryan2003estimating}, myopic sequential design of experiments (SDOE)~\cite{hennig2012entropy,pandita2019bayesian}. and adaptive refinement methods for surrogate modeling\cite{lam2016bayesian,bhaduri2018efficient,bhaduri2018stochastic}.
Batch designs for querying the expensive information source, using the so-called \emph{batch Bayesian optimization} have been proposed in recent years~\cite{azimi2010batch,azimi2011dynamic,azimi2012hybrid,gonzalez2016batch,tran2019pbo}, with promising results.

Consistent with the spirit of Bayesian formalism, a methodology to optimally design sequential experiments~\cite{huan2013simulation,huan2014gradient} for general non-Gaussian posterior states, and nonlinear forward models (for dynamically defined black-box simulators) has been recently proposed by Shen et. al. in \cite{shen2021} and by Cheon et. al. \cite{cheon2021new}. 
The authors~\cite{shen2021} train a design policy, using policy gradient methods of reinforcement learning, that accounts for both feedback and look-ahead design considerations . 
Nonetheless, a major challenge in sequential Bayesian optimal experimental design methods continues to lie in the practical realm of querying multiple experiments in a single shot of a sequence of experiments.

This task, commonly known as batch sampling~\cite{viana2012sequential,azimi2010batch,azimi2011dynamic,tran2019pbo}, is motivated by the need of many challenging engineering problems where the overall budget and experimental logistics solicit more than one different inputs. 
Furthermore, from a theoretical standpoint, the existence of an optimal experimental design for a scenario of a discretized design space with sub-linear relationship between total sample size and design space dimension is unproven. 
Approximate optimal designs have been shown to exist and have been found for a linear relationship \cite{allen2017}. 
This points to the imperative of a monotonic relationship been design space dimension and total sample size when one is seeking optimal experimental design. 
Consequently, large dimensional engineering problems naturally demand extensive design space sampling that would make batch sampling efficient and cost effective.

Extensions of the classical BGO approach, have shown promise, in alleviating the drawbacks of sub-optimal and myopic sampling.
However, the major limitation of these methods that restrict their use in the aforementioned  context are as follows:
a) these methods require a minimum initial number of DOE points in order to train a reliable surrogate model, usually a GP, to being the BGO, and b) the information acquisitions strategies used by these methods do not always account for the long-term expected plausible gains that a batch of points possesses.
In addition to these, the state space in SDOE is a collection of points in the continuous domain of the function resulting in a large state domain.
Hence, making the problem well-suited for deep RL methods.

Reinforcement learning (RL)~\cite{sutton} has been finding wide application in various fields. 
Molecular optimization, steering optimization in robots, job-shop scheduling, and \emph{meta learning} to minimize loss functions in machine learning tasks like image classification are some of the areas where an RL-based approach has yielded promising results~\cite{Zhou_2019, robot, andriotis2020deep, li2017learning}. 
A major reason behind its rising popularity across different disciplines being recent advancements in computational resources, complemented by RL's ability to train agents to perform tasks in unforeseen environments.
These tasks, that are usually deemed hard to  implement and optimize, are characterized by large state domains such as those found in the design of experiments, noisy and sparse observation data, and lack of a model of the environment that can be queried easily. 
The latter of these reasons make such tasks well-suited to be treated with deep RL-based ~\cite{li2017deep} algorithms.

The precedent for using RL methods has been set for the two main aspects covered in this work. 
Optimizing black-box functions and using GP regression as an agent has been done in \cite{li2017learning} and \cite{deisenroth2011pilco, deisenroth2011learning} respectively. 
The task of constructing optimization algorithms with neural networks\cite{bhaduri2021efficient,bhaduri2021stress,bhaduri2020usefulness} that rival state of the art methods such as Adaptive Moment Estimation (Adam) using RL has been accomplished \cite{li2017learning}. 
where the authors use the guided policy search RL method for searching in large sets of involved potential policies. 
The use of model-based RL~\cite{deisenroth2011pilco} with Gaussian process (GP) models that simulate the system dynamics has been researched at length and applied to multiple applications in the fields of optimal control~\cite{deisenroth2010efficient} and robotics~\cite{deisenroth2011learning}. 
In this paper we combine the use of deep RL for these two aspects for optimal batch sampling in SDOE.

Towards this, we explore the use of deep RL~\cite{arulkumaran2017brief} for sequentially sampling from a black-box information source.
The focus here, is on optimizing black-box functions, by selecting multiple experiments to query the information source, at each step of the sequence. 
To this end, we combine RL with Gaussian process (GP) regression~\cite{rasmussen2003gaussian,rasmussen2003gaussianrf} to enhance the state-of-the-art in sequential Bayesian design of experiments. 
We aim to do this with long-term rewards in the formulation of the utility function, that is used to select the batch of experiments.

In other words, we want our algorithm to be non-myopic at each sampling step and to optimize over the entire budget of the expensive function querying.
We assume that a finite budget on the number of experiments or simulations exists and is known at the beginning of the algorithm.
This finite number of simulations allows us to decide the length of the \emph{task}.
The number of sequential queries is encompassed in what is termed as an \emph{episode}.

We formulate a strategy where we train the RL agent on a \emph{training function} in order to learn the so-called \emph{optimal policy}. 
The optimal policy is the batch sequential sampling rule that gives us the globally maximum possible difference in the minimum value for our probabilistic belief of the target function at each sampling step. 
In other words its the policy that gives the best possible return for our budgeted batch sequential sampling. 
The \emph{state} at each step of the task is defined by the previously selected batch of experiments and the statistics of the probabilistic representation of the underlying physical response as a function of the inputs.
This probabilistic representation is formulated using Gaussian process regression.
Given a \emph{state}, the optimal policy prescribes an \emph{action} that is to propose a batch of inputs to query the information source. 
The policy is learnt as a deep neural network~\cite{goodfellow2016deep}, while being updated using standard policy-gradient rules at regular intervals.
Finally, we repeat the process of training the RL policy over multiple iterations, for the \emph{training function} or \emph{training task}.
Once trained, the policy is used to propose batches of experiments for the function of interest, also called \emph{test function} or \emph{unseen task}.
We call our methodology Bayesian Sequential Sampling via Reinforcement Learning (BSSRL).

Our main contributions are as follows:
a) formulating batch sampling for black-box optimization as a task for an RL agent , 
b) using a parametric policy gradient RL method and nearest neighbor sampling to achieve the aforementioned task,
and, 
c) comparing our results with baseline methods for batch sampling using Bayesian optimization.

The outline of the paper is as follows: In Sec. \ref{sec:metho} we describe the mathematical details of the algorithm developed.
Sec. \ref{sec:numerics} demonstrates the performance of the methodology on various numerical experiments and comparison with state-of-the-art methods.
In the last section, Sec. \ref{sec:conclusions}, we discuss where our method has potential to outshine previous methods and possibilities for future work.

\section{Methodology}
\label{sec:metho}
We start with a design space $\mathbf{x}=\{\mathbf{x}_1,\cdots,\mathbf{x}_n\}$ in $\mathcal{R}^d$ and an objective training function $f(\mathbf{x})$. 

\subsection{Gaussian Process Regression}
\label{sec:gpr}

We quantify our belief about the unknown function, probabilistically, using Gaussian process (GP) regression~\cite{williams2006gaussian}. 
A  GP is a series of random variables, all of whose subsets form a joint Gaussian distribution. 
A GPR over a function is a regression for a set of points in the function's domain that is defined at each point in that domain with a mean and standard deviation of a Gaussian random variable. 
Let us present the details of the GPR construction for our unknown function $f$ next.  

The prior GP is given by:
\begin{equation}
\label{eqn:prior}
\begin{aligned}
    f \sim GP(0, k(\mathbf{x},\mathbf{x}'))
\end{aligned}
\end{equation}

where,

\begin{equation}
    \label{eqn:cov_kern}
    k(\mathbf{x},\mathbf{x}') =  {s^2}\exp \left\{ { - \frac{1}{2}\sum\limits_{j = 1}^d}{\frac{{{{({x_j} - {x_j}')}^2}}}{{\ell_j^2}}} \right\}.
\end{equation}

Defining this covariance and mean function (considered zero in this work), allows one to fully specify a prior probabilistic belief on the space of functions.
More details on the structure of covariance functions used in Gaussian processes can be found in~\cite{williams2006gaussian}.
The covariance function defined in (\ref{eqn:cov_kern}) encodes our prior beliefs about the smoothness and magnitude of the response with the parameters $\ell_{j}>0$ and $s^2$. 
The symbol $\ell_{j}>0$ in  (\ref{eqn:cov_kern}) is the lengthscale of the $j$-dimension of the input space.
This parameter quantifies the correlation between the function values at two different inputs.
The $s^2$ in (\ref{eqn:cov_kern}) is the signal strength of the GP representing the scale of this unknown function.
Together, these parameters are known as the \emph{hyper-parameters} of the covariance function. 
We will denote these hyper-parameters by $\btheta$.

In order to condition this prior belief on $f$, on the available data that is usually obtained by querying the expensive function, we model the set of observations as independent Gaussian measurements with a finite variance.
We consider a sample of the initial set of points $\hat{\mathbf{x}}=\{\hat{\mathbf{x}}_1,, \cdots,\hat{\mathbf{x}}_n\}$ in the domain $\mathbf{R}^{d}$, where $d$ stands for the dimensionality of the input domain. 
We specify the sampling method in section \ref{sec:sdoe}. 
We then find the corresponding single-valued outputs for these sampled points $\hat{\mathbf{y}}=\{\hat{y}_1,\hat{y}_2,...,\hat{y}_n\}=\{f(\hat{\mathbf{x}}_1),f(\hat{\mathbf{x}}_2),...,f(\hat{\mathbf{x}}_n)\}$. 
We model this, so-called likelihood probability distribution, for the input-output pairs, is mathematically expressed as follows:

\begin{equation}
\label{eqn:likelihood}
    \begin{array}{ccc}

    \hat{y}_1 \sim \mathcal{N}(f, \sigma^{2})  \nonumber \\ 
    
    p(\hat{\mathbf{y}}) = \prod\limits_{i=1}^{n}p(\hat{y}_1),
    
    \end{array}
\end{equation}

We use the likelihood and the prior GP, and leverage Bayes' rule to derive the posterior state of belief for $f$.
\begin{equation}
\label{eqn:posterior}
\begin{aligned}
    f| \mathbf{y}_n, \mathbf{x}_{n}  \sim GP(m_{n}, k_{n}),
\end{aligned}
\end{equation}

In Eq. \ref{eqn:posterior}, $m_{n}$ and $k_{n}$ are the posterior mean and covariance functions for the posterior state of belief on $f$ which is also a GP and are defined as follows:

\begin{equation}
    \label{eqn:posterior_mean}
    m_n(\bx) = \left(\bk_n(\bx)\right)^{T}\left(\bK_n + \sigma^2\bI_n\right)^{-1}\bY_{n},
\end{equation}



\begin{equation}
    \label{eqn:predictive_covariance}
    k_n(\bx,\bx') = k(\bx, \bx') - \left(\bk_{n}(\bx)\right)^{T}\left(\bK_n + \sigma^2\bI_n\right)^{-1} \bk_n(\bx'),
\end{equation}

with,
$\bk_n(\bx) = \left(k(\bx,\bx_1), \dots, k(\bx,\bx_n)\right)^T$, is the \emph{posterior covariance} function.

In particular, at an untried design point $\tilde{\bx}$ the point-predictive posterior probability density  of $\tilde{y} = f(\tilde{\bx})$ conditioned on the hyperparameters is:
\begin{equation}
    \label{eqn:point_predictive}
    p(\tilde{y}|\tilde{\bx}, \hat{\mathbf{x}}_n, \hat{\mathbf{y}}_n, \btheta) = \mathcal{N}\left(\tilde{y}\middle|m_n(\tilde{\bx};\btheta), \sigma_n^2(\tilde{\bx};\btheta)\right),
\end{equation}

where,
$\sigma_n^2(\tilde{\bx};\btheta) = k_n(\tilde{\bx},\tilde{\bx};\btheta)$.

In this work, we leverage the implementation of GP regression, via the open-source \emph{sklearn} library in the Python programming language.
In order to speed-up the computation, we estimate $\btheta$ as the solution that maximizes the likelihood of observing the training data. 
We omit the fully Bayesian formalism in inferring the hyperparameters of this intermediate GP while acknowledging the potential drawbacks~\cite{gelman1995bayesian,bilionis2013multi} of the same. 
The estimated hyperparameters of the GP allow us to sample from the posterior predictive distribution of $f$, with a predictive mean and uncertainty at unobserved points in the design space.

\subsection{Sequential design of experiments}
\label{sec:sdoe}
For the sequential design of experiments (SDOE) we  sample multiple batches for an episode of batch sequential sampling and sequentially reform the GP model for the fitting problem until we reached our sampling budget. 
During each episode step in said episode, we face the challenge of picking a batch of points in the function domain that help us find the function's best possible minimal value for our fixed budget of sampling batch size and number of batches. 

We denote the array of batches sampled from the objective function as follows:
\begin{equation}
    \label{eqn:sampled_points}
\mathcal{X}=(\{\mathbf{\bar{x}}^{(1)}_{1},...,\mathbf{\bar{x}}^{(1)}_{n}\},...,\{\mathbf{\bar{x}}^{(m)}_{1},...,\mathbf{\bar{x}}^{(m)}_{n}\})
\end{equation}

where $m$ is the number of batches we are able to sample and $n$ is the fixed cardinality of each batch. We will set $S=mn$ to be the total number of sampled points. 
Let the location, in the function domain, of the minimum of the final GP  be given by: 

\begin{equation}
    \label{eqn:argmin}
    \mathbf{\breve{x}}=\argmin\limits_{\mathbf{x}} \mathbb{E}[y^{(m)}](\mathcal{X}),
\end{equation}

where $\mathbb{E}[y^{(m)}]$ is the mean of the final GPR achieved after finishing the process of batch sequential sampling. 

The final objective of our algorithm is to find:
\begin{equation}
    \label{eqn:Objective1}
   \min\limits_{\mathcal{X}} f(\mathbf{\breve{x}}(\mathcal{X}))
\end{equation}


Note that $\mathbf{\breve{x}}$ is a dependent variable of $\mathcal{X}$ and that the minimization problem in (\ref{eqn:Objective1}) is not for the GPR but for the true function's global minimum. 
Hence, the overall problem is a composite optimization problem with the composed functions being the objective function and the function relating the sampled batches of points to the location of the minimum of the GPR's posterior. 
The latter function is modeled as discussed in section \ref{sec:gpr}.

As can be gathered from the formulation, this task of sequentially designing the experiments has practical limitations. 
The main limitations are the budgeted evaluation of the objective function carried out sequentially using GPR and consequently the inaccessibility of the objective function's gradient.

\subsection{System dynamics for sequential optimal design of experiments}
\label{sec:rl_sdoe}
Reinforcement learning is a decision making method that that has a learner (typically referred to as an agent) observe, partially or completely, the state in a given environment, react to it, and get feedback associated with the reaction usually in the form of a reward or cost. 
The learning happens when the agent maximizes the cumulative reward or minimizes the cumulative cost. 
The accumulation function is designed with the timing requirements on the reward/cost in mind. 
We leverage similar ideas as Li. et al. ~\cite{li2016learning}, where the authors learn a policy to optimize unseen functions, to define the state and action for our problem.
Formally, the reinforcement learning system dynamics for sequential sampling can be modeled as follows:

\begin{enumerate}
    \item \textbf{State($S_i$)}: The sample of domain points chosen $X^{(i)}=\{x^{(i)}_1,x^{(i)}_2,...,x^{(i)}_n\}$ and the GPR statistics. 
    The GPR statistics are the minimum value's mean ($m^{(i)}_n(\tilde{\bx}_{\min})$) and standard deviation ($\sigma_n^{(i)}(\tilde{\bx}_{\min})$) at episode step   $i$. Here $\tilde{\bx}_{\min}$ is the argument of the minimum point in the GPR. 
    \item \textbf{Action($A_i$)}: The agent chooses the new sample $X^{(i+1)}$ in the design space following a policy. 
    The policy $\pi(a|s)$ gives the probability of taking action a given state s. 
    The action involves the following three steps: 
    a) choosing the parameter's for the sampling distribution according to the policy, b) sampling the points from said distribution, and c) adding the sample (or their nearest neighbors if we are on a grid) to our repository of points to condition the  GP. 
    The action is only the first of the above three steps and the later two steps are how it guides the state transition. 
    \item \textbf{Reward($r_i$)}: The reward linearly combines the classical exploration and exploitation priorities in optimal experimental design. 
    The reward function is given by 
    \begin{equation}
    \label{reward_function}
    r_i=-(m^{(i)}_n(\tilde{\bx}_{\min})-m^{(i-1)}_n(\tilde{\bx}_{\min})+ \alpha(\sigma_n^{(i)}(\tilde{\bx}_{\min})-\sigma_n^{(i-1)}(\tilde{\bx}_{\min})))
    \end{equation}
    Here $\alpha$ is the preference parameter between exploration and exploitation in the design space.
    
    We use the this linear form as it allows to incorporate into the reward both the exploration and the exploitation effects.
    One can choose other forms for the reward function based on their use case.
    Thus, making the methodology applicable to other inference tasks like inferring the expectation or learning extreme statistics of the underlying function.
\end{enumerate}

The transition probability distribution for mapping how an action produces a new state given the previous state (sampling a new batch leads to new GPR statistics given the previous sample and GPR statistics) is not used. 
This distribution is intractable in our setup since we do not include the complete information of the GPR in our state variable as we cover the episode step  s of the RL episode. 
Hence, the RL approach described above, named BSSRL, is model-free.
This will help with avoiding over-fitting to training function geometry~\cite{li2016learning}. 

We compute the cumulative reward as 
   \begin{equation}
    \label{cumulative_reward_function}
R=\sum\limits_{i=1}^{N} \gamma^i r_i
\end{equation}
where $\gamma$ is called the discount factor and is between 0 and 1. 
When $\gamma$ is small earlier rewards in an episode are prioritized. 
Note that when $\gamma=1$ the reward sum telescopes and leaves only the difference of the initial and final GPR statistics. 
This depicts that the objective will not be myopic.

We notice that the system evolution is a partially observable Markov decision process. 
The state only contains partial information needed to have a decision process with the Markov property. 
Given the state and action of a given step $i$ in an episode the distribution for the next state in step $i+1$ is not the same if we knew the points from all the previous steps ($\{1,2,...,i-1\}$) or not. 
This inhibits us from resolving the system using dynamic programming methods that use Bellman's equation and by extension the Markov property \cite{sutton}. 
Instead we use REINFORCE~\cite{williams1992simple}, the standard policy gradient method. 
The method we use is explained next. 

\textbf{Continuous policy with policy gradient}:
This method uses a normally distributed policy distribution with learn-able parameters. 
Because the policy is differentiable we can use policy gradient RL methods to learn the optimal policy for batch sequential sampling for a class of continuous objective functions. 
When our policy model is not differentiable or when the objective is not a continuous function but rather a discrete data set on a mesh-grid we can use \say{nearest neighbor} approach to modify our policy gradient RL method to overcome the challenge. 
The policy gradient RL method we use throughout is REINFORCE~\cite{williams1992simple}, wherein we use a neural network (DNN) to parameterize the policy. 
In this work, the architecture of the neural network is chosen to be fixed across all experiments, and we use a standard two-layer multi-layer perceptron DNN with 16 nodes each and ReLU~\cite{goodfellow2016deep}  activations.

The algorithm is given below:
\begin{algorithm} [H]
\caption{\textbf{Bayesian Sequential Sampling with Reinforcement Learning}}
\begin{algorithmic}
\item Initialize distribution for the objective training function class
\item Determine number of training functions ($s$)
\item \textbf{While} $\mathbf{i<s+1}$ \textbf{do}
   \item\par \hspace{0.5cm}  Generate sample function $f_i$ from the class of objective functions
    \item\par \hspace{0.5cm} Initialize mesh, episode size (m), number of episodes (t)
    \item\par \hspace{0.5cm} \textbf{While} $\mathbf{j<t+1}$ \textbf{do}
         \par\par \hspace{0.5cm} Initialize state ($X^{(0)}$ \& GPR statistics ($m_0,k_0$)), total episode reward, and storage
         \par\par \hspace{0.5cm} \textbf{While} $\mathbf{k<m+1}$ \textbf{do}
         \par\par\par \hspace{1cm} Determine policy for the given state by selecting policy parameters
         \par\par\par \hspace{1cm} Take a sample averaged action from the policy distribution and apply it
         \par\par\par \hspace{1cm} Record new state and reward $r_k$ in storage
          \par\par\par \hspace{1cm} Update state 
          \par \par \hspace{0.5cm} \textbf{end while}
         \par\par \hspace{0.5cm} \textbf{While} $\mathbf{k<m+1}$ \textbf{do}
        \par\par\par  \hspace{1cm} Compute cumulative reward from episode step k to m
         \par\par\par \hspace{1cm} Find policy distribution for the state at k
         \par\par\par \hspace{1cm} Update policy wrt. cumulative reward using REINFORCE
         \par \par \hspace{0.5cm} \textbf{end while}
    \par \textbf{end while}
\item \textbf{end while}
\end{algorithmic}
\end{algorithm}

We compute the updates for the parameters of the policy distributions for a given episode ($j$) as follows. 
\begin{equation} \label{mean_policy}
    \begin{aligned}
        &\mu_{i}^{(j)}=\alpha (R[m]-R[j]) (\frac{1}{(\omega_{i}^{(j)})^2})(ac_{i}-\mu_{i}^{(j)}),     \end{aligned}
\end{equation}
\begin{equation}  \label{st_dev_policy}
    \begin{aligned}
        &(\omega^{(j)}_{i})^2= -0.5\alpha (R[m]-R[j])((\frac{1}{(\omega_{i}^{(j)})^2}) (ac_{i} - \mu_{i}^{(j)})^2+(\frac{1}{(\omega_{i}^{(j)})^2}))
    \end{aligned}
\end{equation}
Here $\mu_i^{(j)}$ and $(\omega_{i}^{(j)})^2$ are the mean and standard deviation of the policy distribution at the ith step of the episode. 
$ac_i$ is the action taken by the agent for the state at episode step i and $\alpha$ is the policy learning rate. 
The neural net takes as inputs, the state of the system (i.e. $X^{(i)}$ and $(m_i,k_i)$) and maps them to the policy's statistical parameters. 
This is done like dynamic supervised learning where the target function (the policy parameter generator) gets redefined with new data points (updated parameters due to reward signals) in each nested period in each episode. 

We note that the BSSRL scheme is a meta learning method. 
The idea of meta learning with RL for optimization is discussed in \cite{li2016learning}. 
There we see that an RL agent's trained scheme for finding the global optimum point of a given function is an alternative to using schemes such as BFGS. 
In such popular schemes we have a systemically and thoroughly defined strategy of accessing a function's value and gradient at a given point and deciding on the next coordinates to access those values. 
With BSSRL however we make the agent train and gather its own meta lessons and adopt a strategy about how to go about using the state to decide on the next sample to select irrespective of the function at hand. 
This strategy can not be a gradient based method such as steepest descent because the agent will have access to the function value at the points it samples but will be limited from knowing the gradient. 
Instead it will have a surrogate for the gradient through the GPR minimum's statistics and the batch sample's function values. 
Using this information in the state the BSSRL agent constructs an optimization strategy through the reward function. 
Asking where the training of the object does not apply (what is the boundary of the set of functions it applies on) is akin to asking for what functions method of steepest descent is ineffective. 
In the case of the steepest descent we have an exact answer for this question since we know exactly what the overall strategy there is. However, for BSSRL we do not know when it would fail since we do not know what empirical strategies it picks up from its training. 
In Fig. \ref{fig:airfoil_RL_results_1} we see BSSRL doing worse that Batch BO for large batch sizes because the policy's input space (domain) has dimension $d \times n+2$. 
This linear growth makes it hard to train the policy well with a fixed budget of episodes.

If the objective function is a black-box function on a mesh-grid, a predefined grid of inputs at which experiments or simulations can be performed, we add an extra step in the training and testing. 
First, we take multiple batches sampling from the distribution with policy given parameters and average them to get a single batch. 
Then we find the nearest neighbor in the predefined grid of inputs, for each point in the averaged batch while consistently removing points from the grid file that are already chosen. 

In each example we have the same domain for the training and testing functions.

\section{Numerical examples}
\label{sec:numerics}
\subsection{Ackley and Booth function}
\label{sec:synthetic}
We consider a problem of minimizing a toy function, namely the \emph{Booth}~\cite{sharma2015comparative} function.
The RL-based policy is learned by training on a different mathematical function, namely the \emph{Ackley} function.
The main motivation for using this set of mathematical functions is that both the functions have similar overall patterns in terms of sharpness of gradients, presence of numerous local minima, and the same number of inputs.
The analytical forms of the \emph{Ackley} and the \emph{Booth} functions are given in Eqs.\ref{eqn:ackley} and \ref{eqn:booth}, respectively.
In order to retain numerical stability, we normalize the inputs and the function outputs to take values between $-3$ and $3$, using standard linear scaling.
We start the batch BO methodology with an initial dataset of three points. 
Since, this is a two-dimensional problem, we start from a low-sample regime in order to better compare the proposed methodology with the batch-BO.

\begin{equation}
\label{eqn:ackley}
\begin{aligned}
    f(\mathbf{x}) = -20\exp^{-0.2 \sqrt{\frac{1}{2} \sum_{i=1}^{2}x_{i}^{2}}}
    - \exp^{\frac{1}{2} \sum_{i=1}^{2}\cos{2\pi x}} + 20 + e,
\end{aligned}
\end{equation}
where $\mathbf{x} \in [-32.768, 32.768]^2$.

\begin{equation}
\label{eqn:booth}
\begin{aligned}
    f(\mathbf{x}) = (x_{1} + 2x_{2} -7)^2 + (2x_{1} +  x_{2} - 5)^2,
\end{aligned}
\end{equation}
where $\mathbf{x} \in [-10, 10]^2$.

In this example, we aim to demonstrate two characteristics of the methodology.
First, the difference in performance on minimizing the Booth function, between the BSSRL 
that is trained for a reasonable number of episodes and a Bayesian optimization (BO) based batch sampling method~\cite{kristensen2019industrial,viana2012sequential}.
Second, the characteristics of the batches of designs that are sampled by the BSSRL methodology and the BO method, for different sizes of the batch.

\begin{figure}[H]
	\centering
	\includegraphics[scale=0.45]{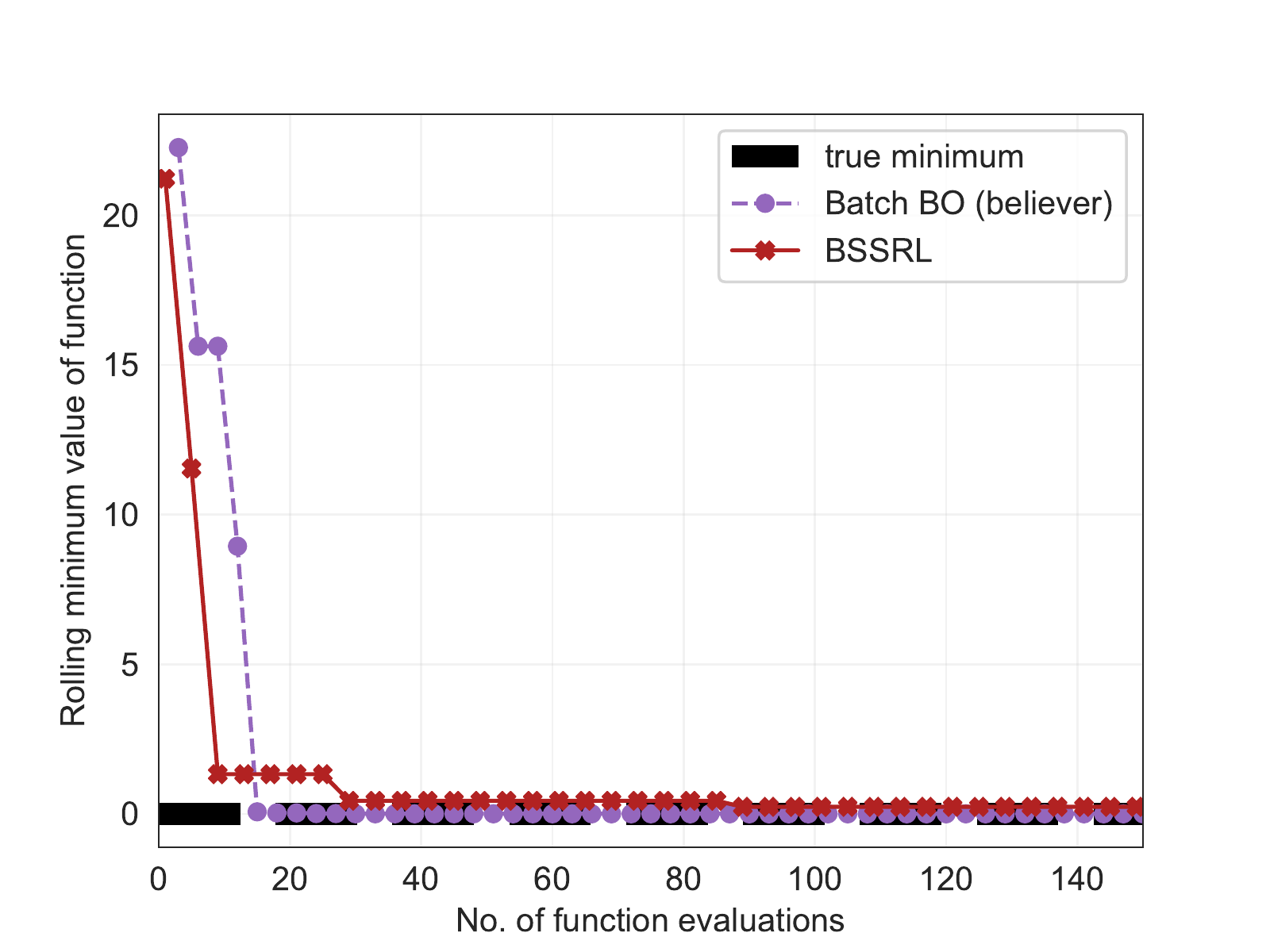}
	\includegraphics[scale=0.45]{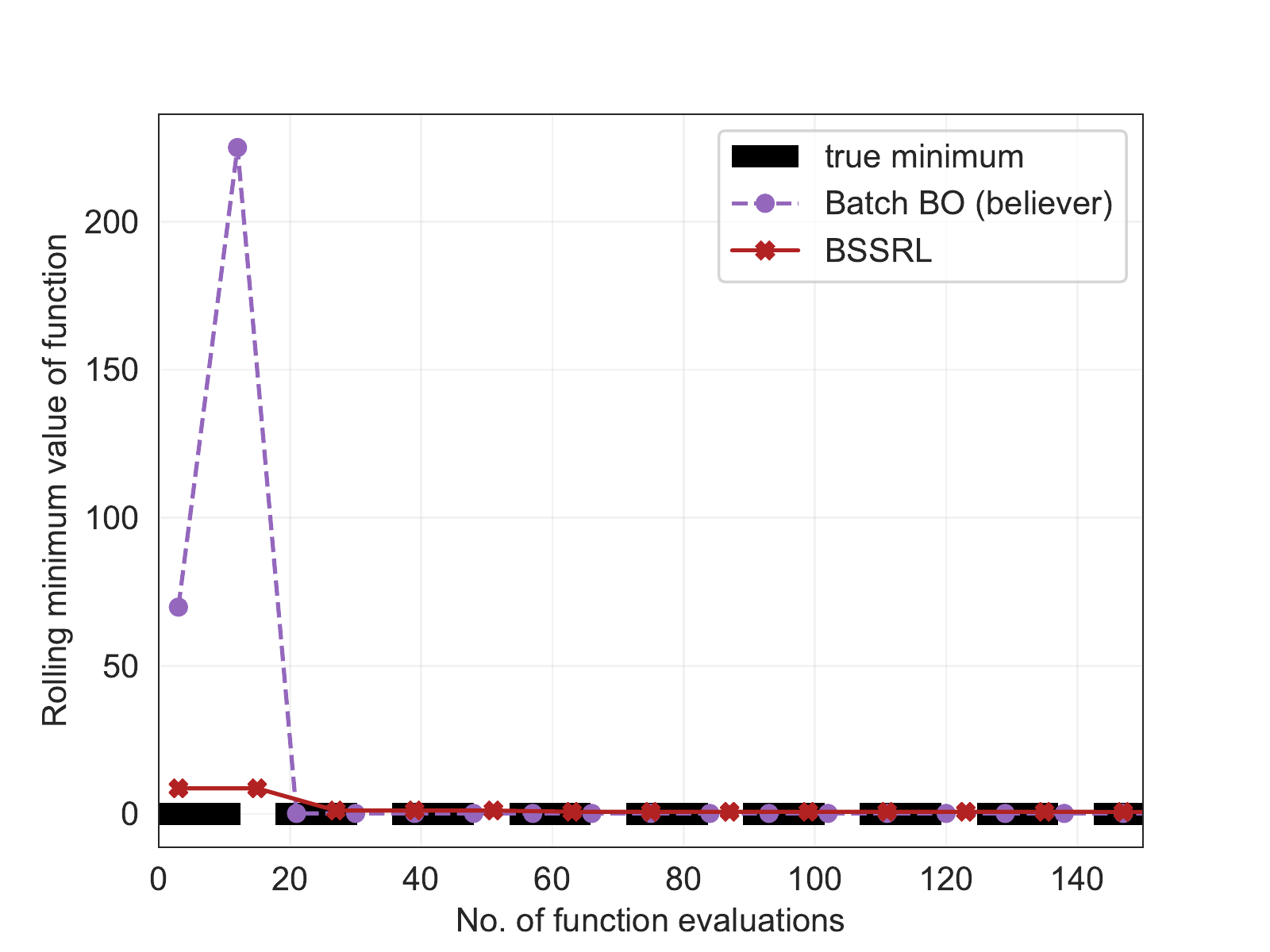}
	\includegraphics[scale=0.45]{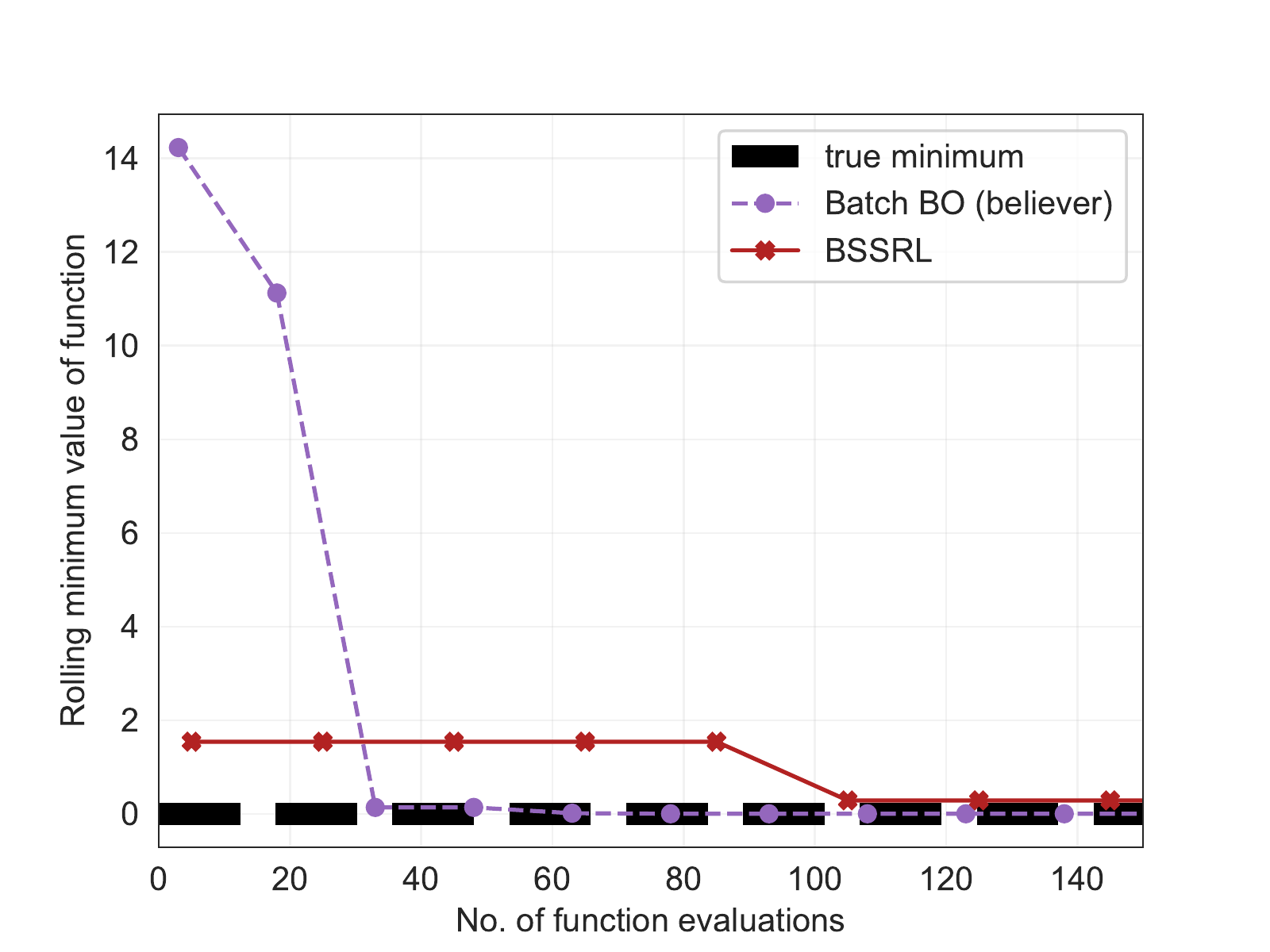}
	\includegraphics[scale=0.45]{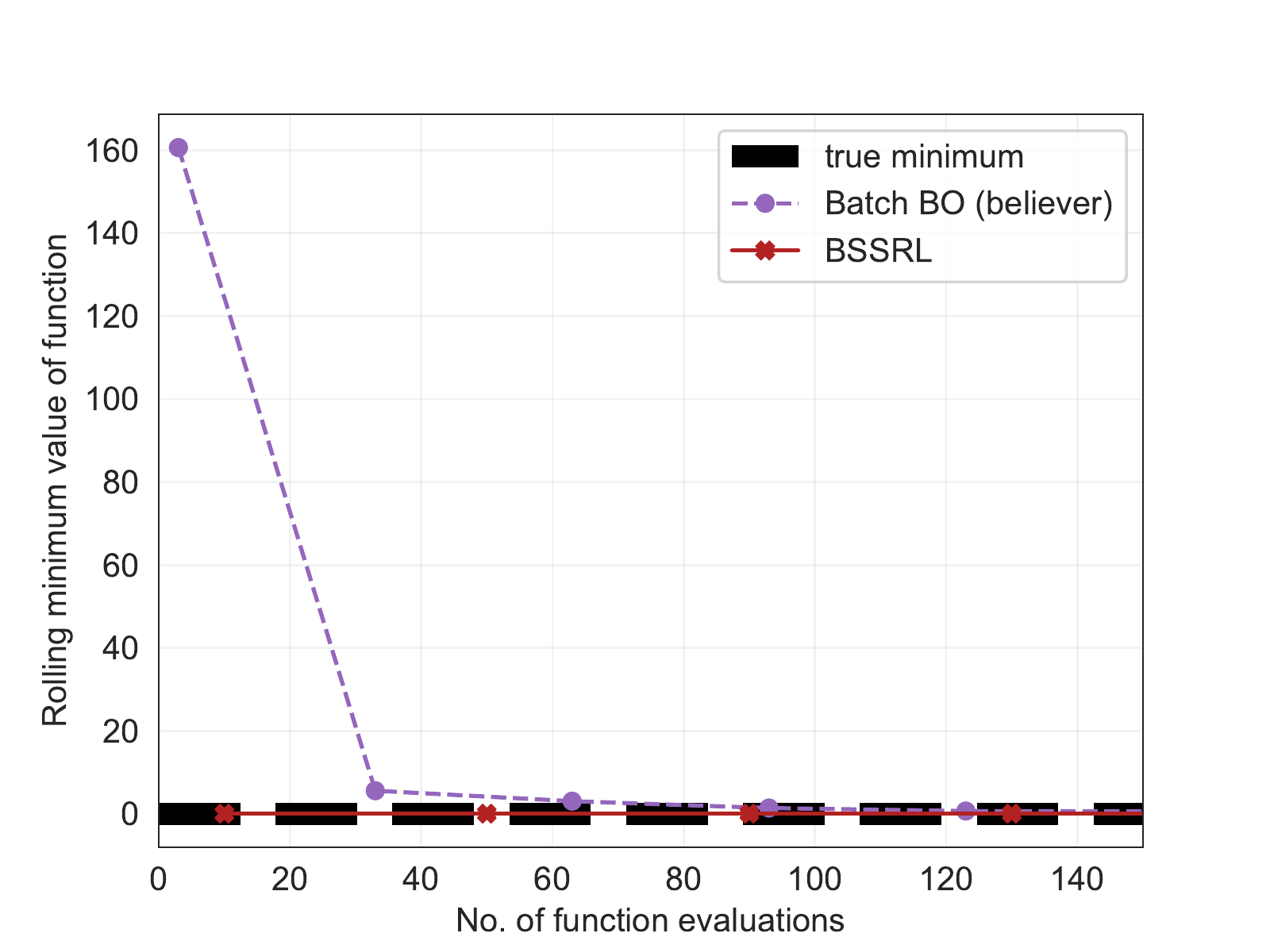}
	\caption{Convergence of the algorithm, trained on Ackley function and tested on the Booth function, with varying batch sizes. 
	Subfigure~(a) single-point per iteration, 
	~(b) 3 points per iteration, ~~(c) 5 points per iteration, and~(d) 10 points per iteration.}
	\label{fig:synthetic_RL_results_1}
\end{figure}

Fig. \ref{fig:synthetic_RL_results_1} shows the difference between the trained BSSRL and the batch BO method, in optimizing the \emph{Booth} function, for different values of batch sizes.
The subfigures show the difference in the number of function evaluations or the number of data required by the two methods to converge to the minimum value of the function.
For smaller batch sizes in Subfigures~(a) and ~(b), it can be seen that the number of data needed are very similar for both the methods.
Whereas with a larger batch size, as seen in subfigures~(c) and ~(d), we see the BSSRL converge faster in comparison with the baseline method.
It reaches the global minimum value in less than 20 data points, whereas the BO based method takes at least thrice as many data points to find the global minimum.

\begin{figure}[H]
	\centering
	\includegraphics[scale=0.45]{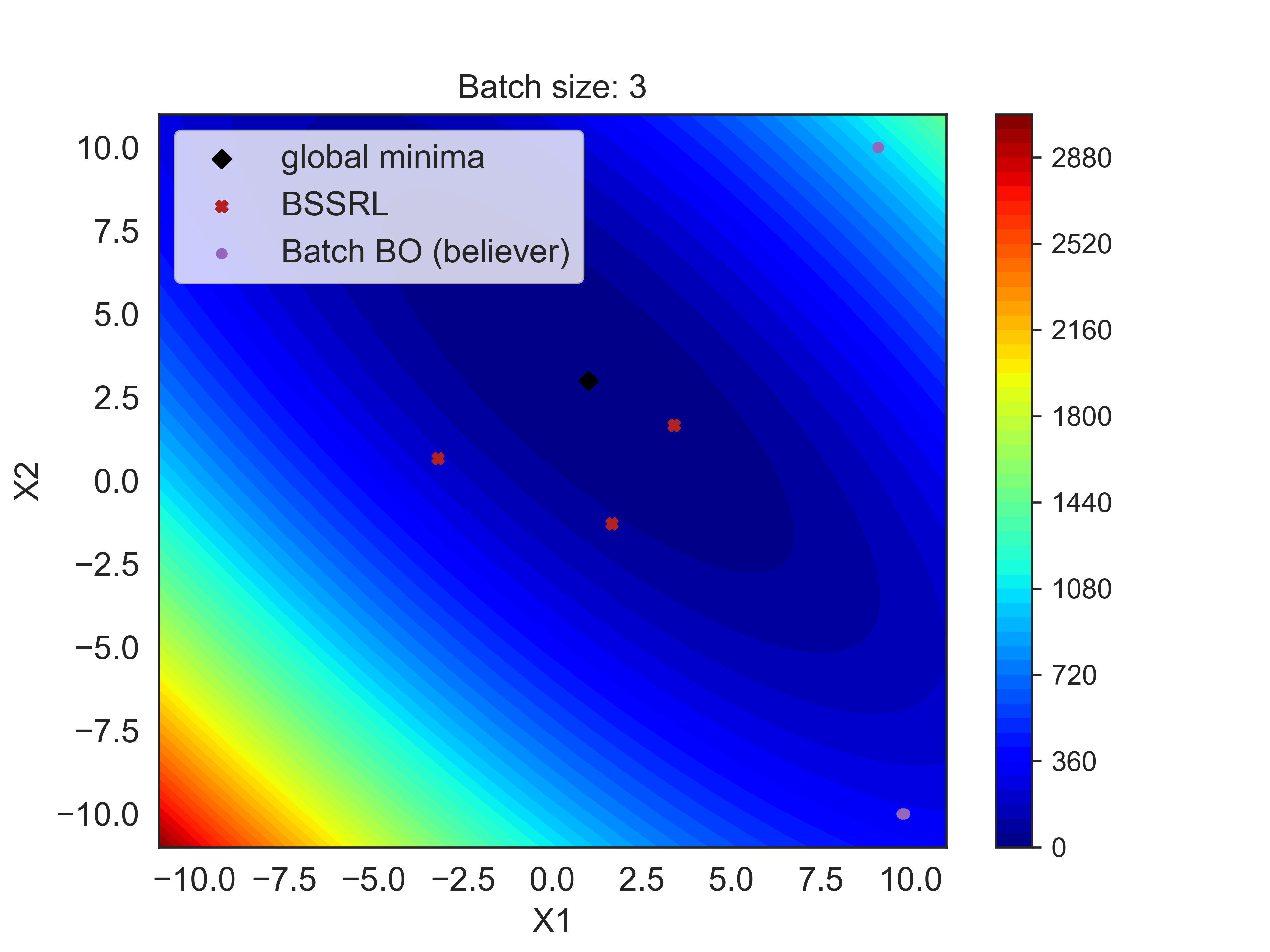}
	\includegraphics[scale=0.45]{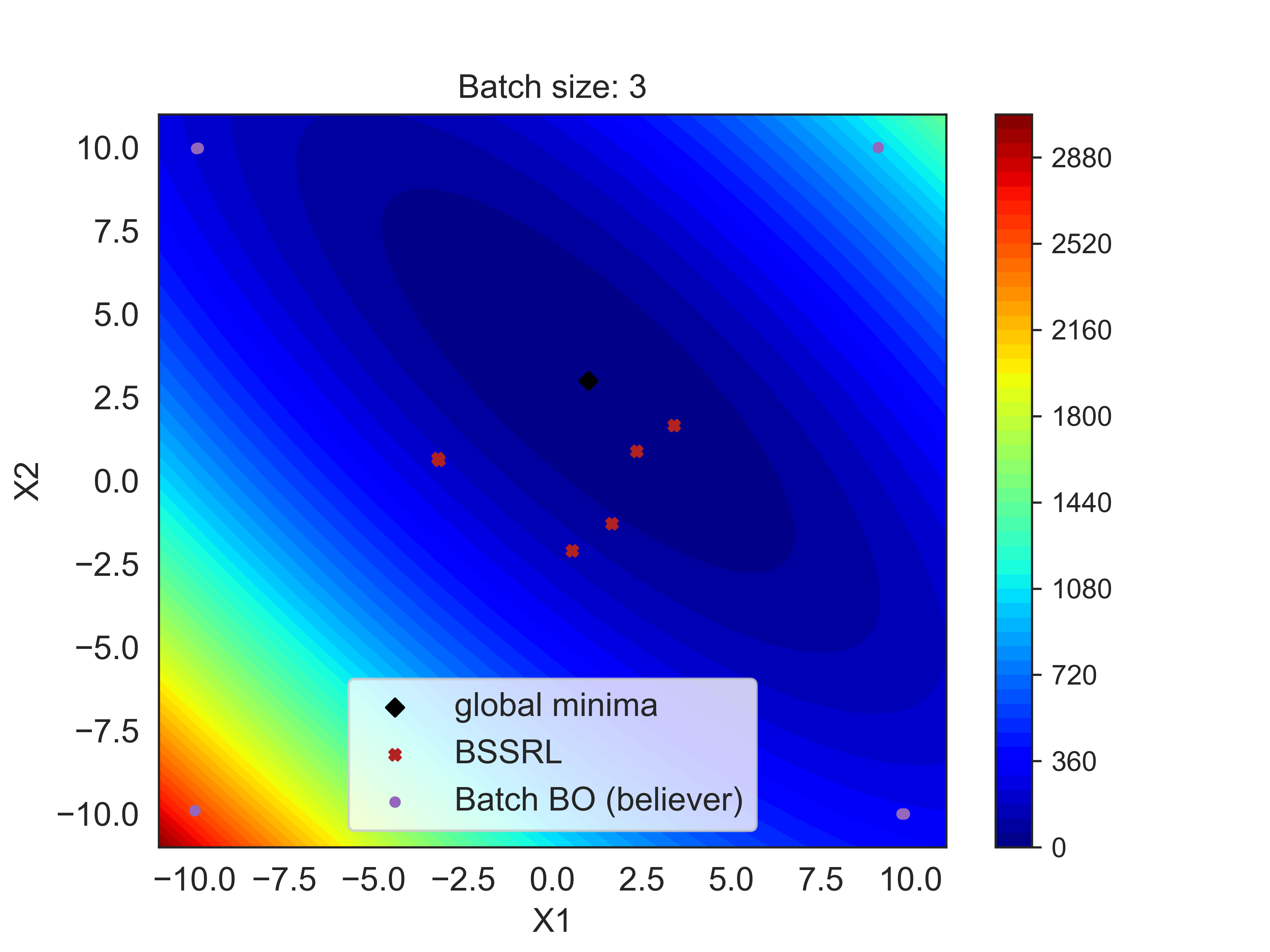}
	\includegraphics[scale=0.45]{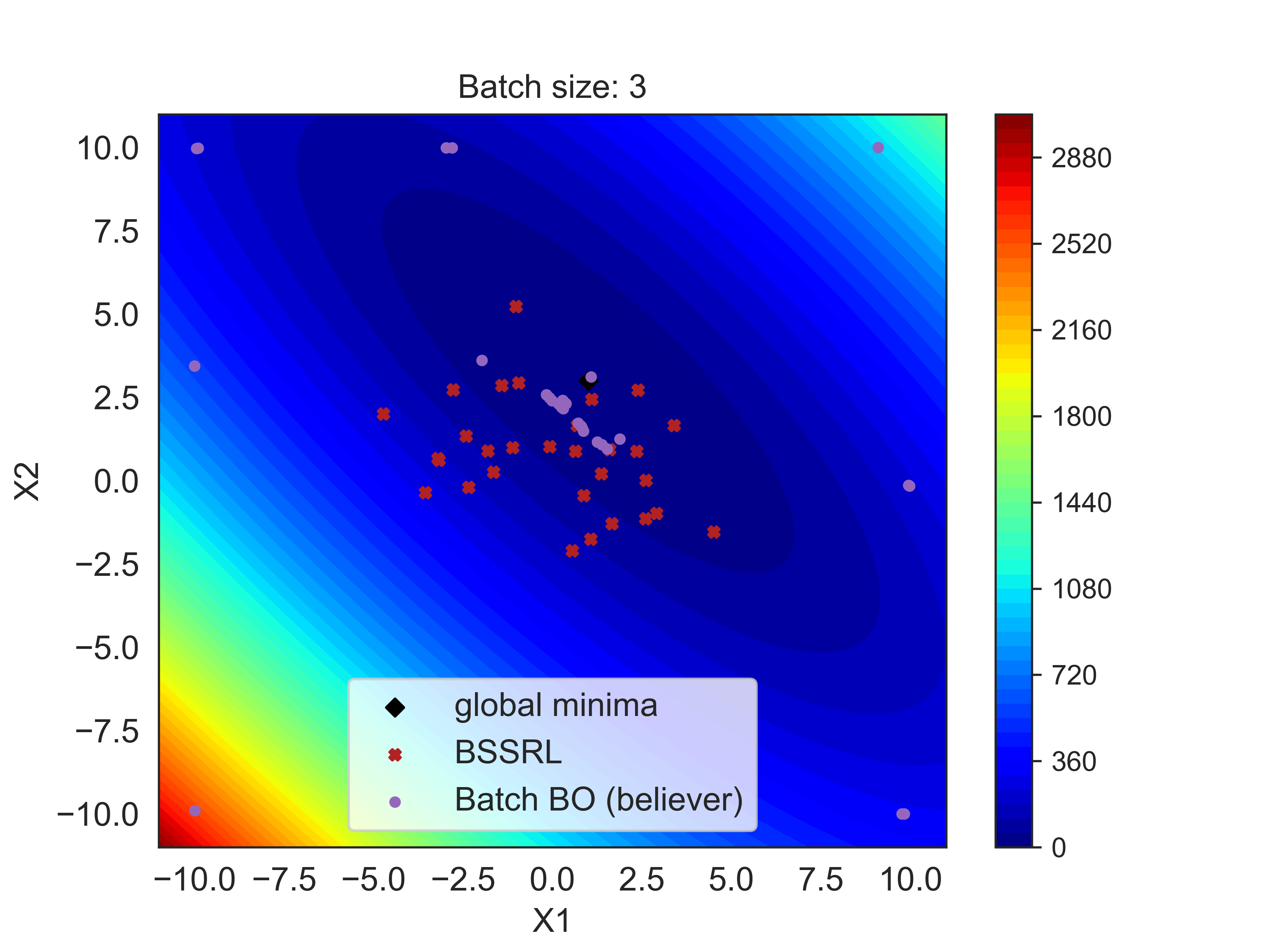}
	\includegraphics[scale=0.45]{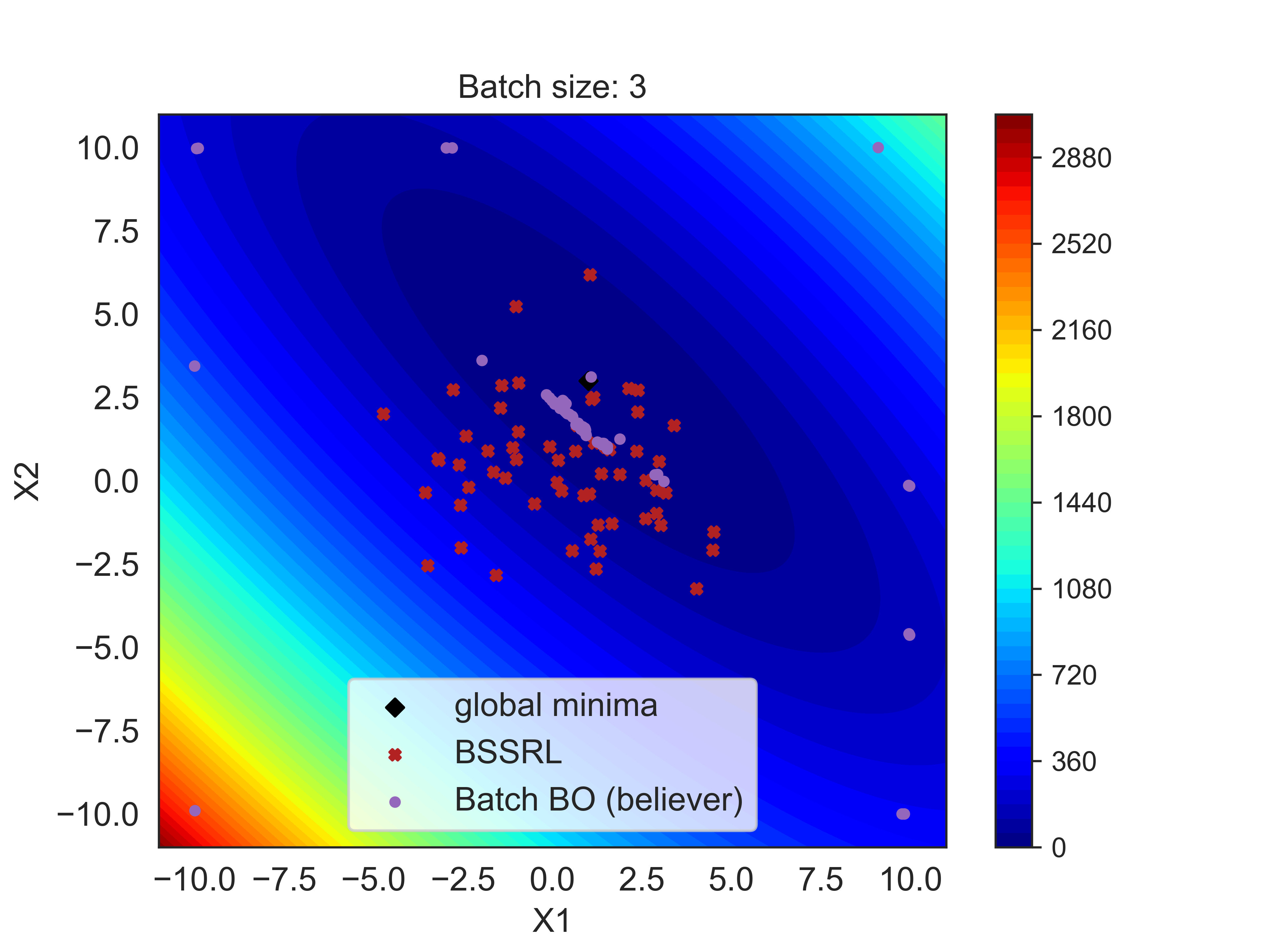}
	\caption{Designs suggested by the BSSRL and the Bayesian optimization with batches of 3 points per batch and 50 total batches (steps).}
	\label{fig:synthetic_RL_results_2}
\end{figure}

In addition Fig. \ref{fig:synthetic_RL_results_2}, the difference in convergence to the minimum value of the \emph{Booth} function, is shown in comparison with the baseline method for the designs or inputs that are selected by each algorithm when the size of the sampled batch is three.

The subfigures in Fig. \ref{fig:synthetic_RL_results_2}, show the performance of the proposed algorithm, for the first, the second and the tenth and the twentieth iteration of the BSSRL and the BO algorithm, in subfigures (a), (b), (c) and (d) respectively.
The global optimum for the \emph{Booth} function is at $\mathbf{x}_{o}=[1, 3]$, with a function value of $f(\mathbf{x}_{o}) = 0$.
In Subfigure~(a), the two methods add three new points, the red crosses for the BSSRL and the purple dots for the BO-based methodology. 
The global optimum is represented by the black diamond.
The subsequent subfigures show how the two methods evolve in evaluating the two-dimensional design space.
The batches of designs selected by the BSSRL constantly explore areas near the global minimum from the very first iteration.
The batch BO method selects points in sub-optimal regions of the design space in the beginning of the sequential process.
Moreover, the points selected by the batch BO are clustered around a specific region which is close to the global optimum.
The BSSRL selects the points that are relatively further apart and appear to be space-filling in the regions around the global minimum.

\begin{figure}[H]
	\centering
	\includegraphics[scale=0.45]{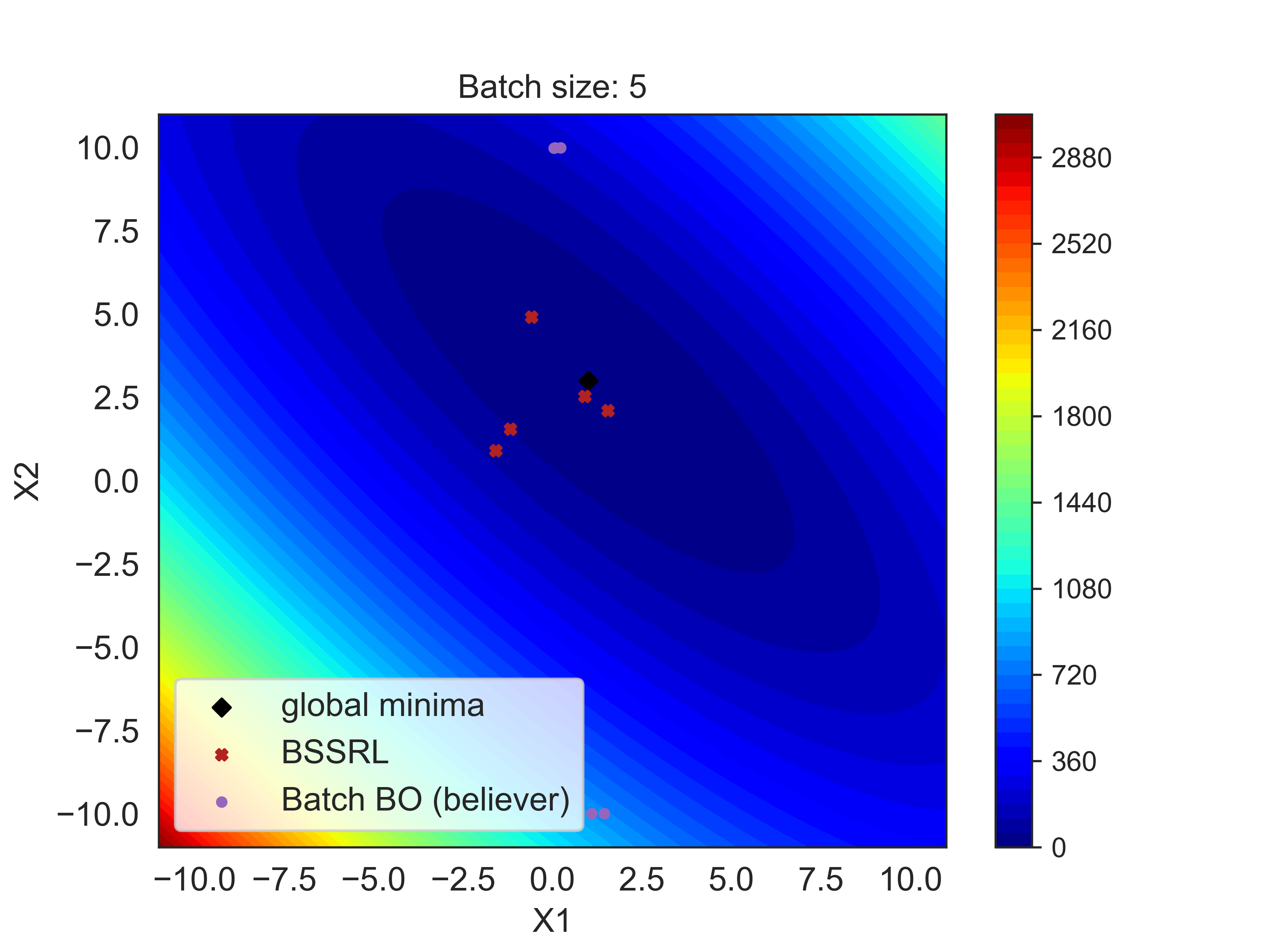}
	\includegraphics[scale=0.45]{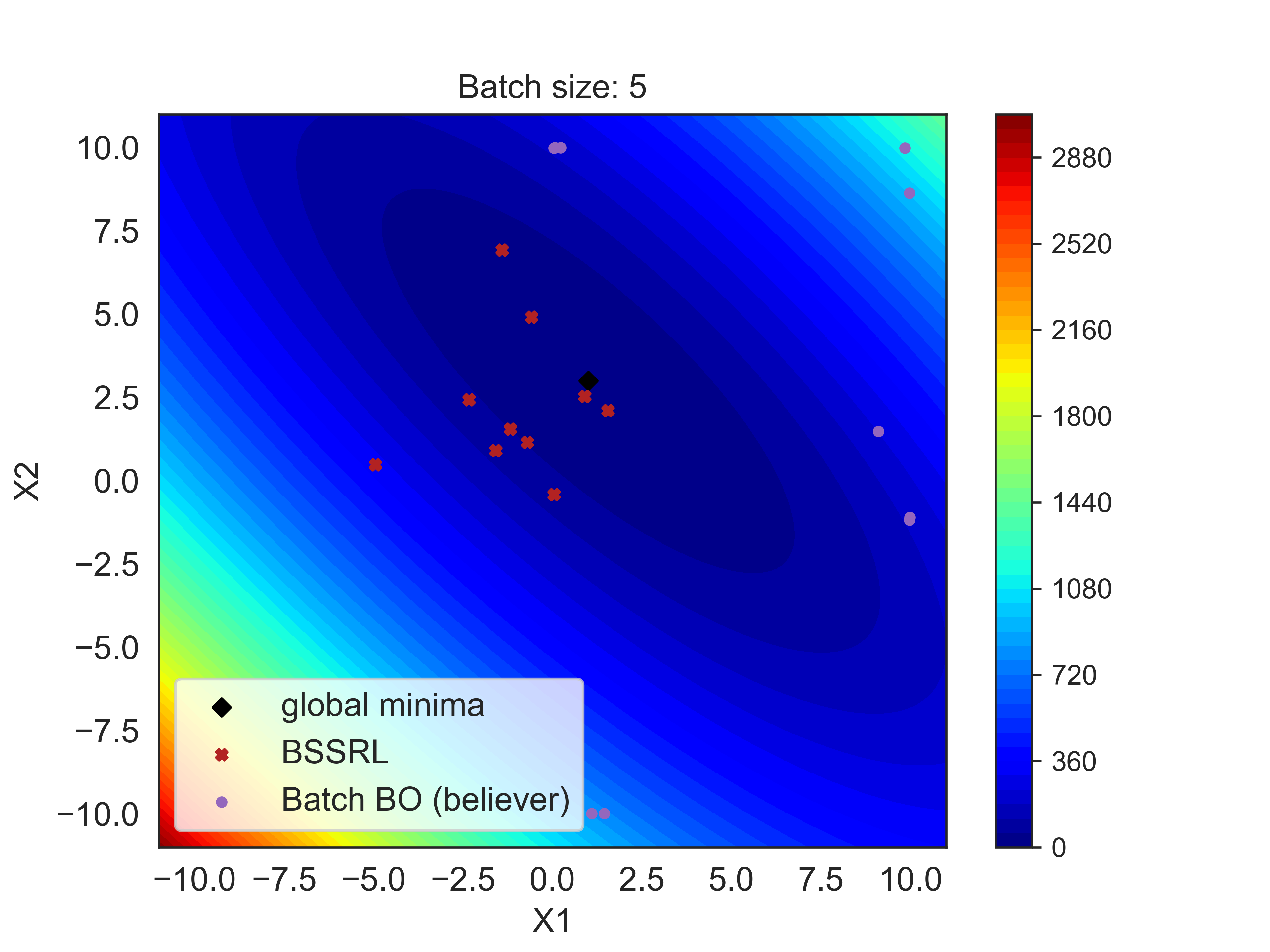}
	\includegraphics[scale=0.45]{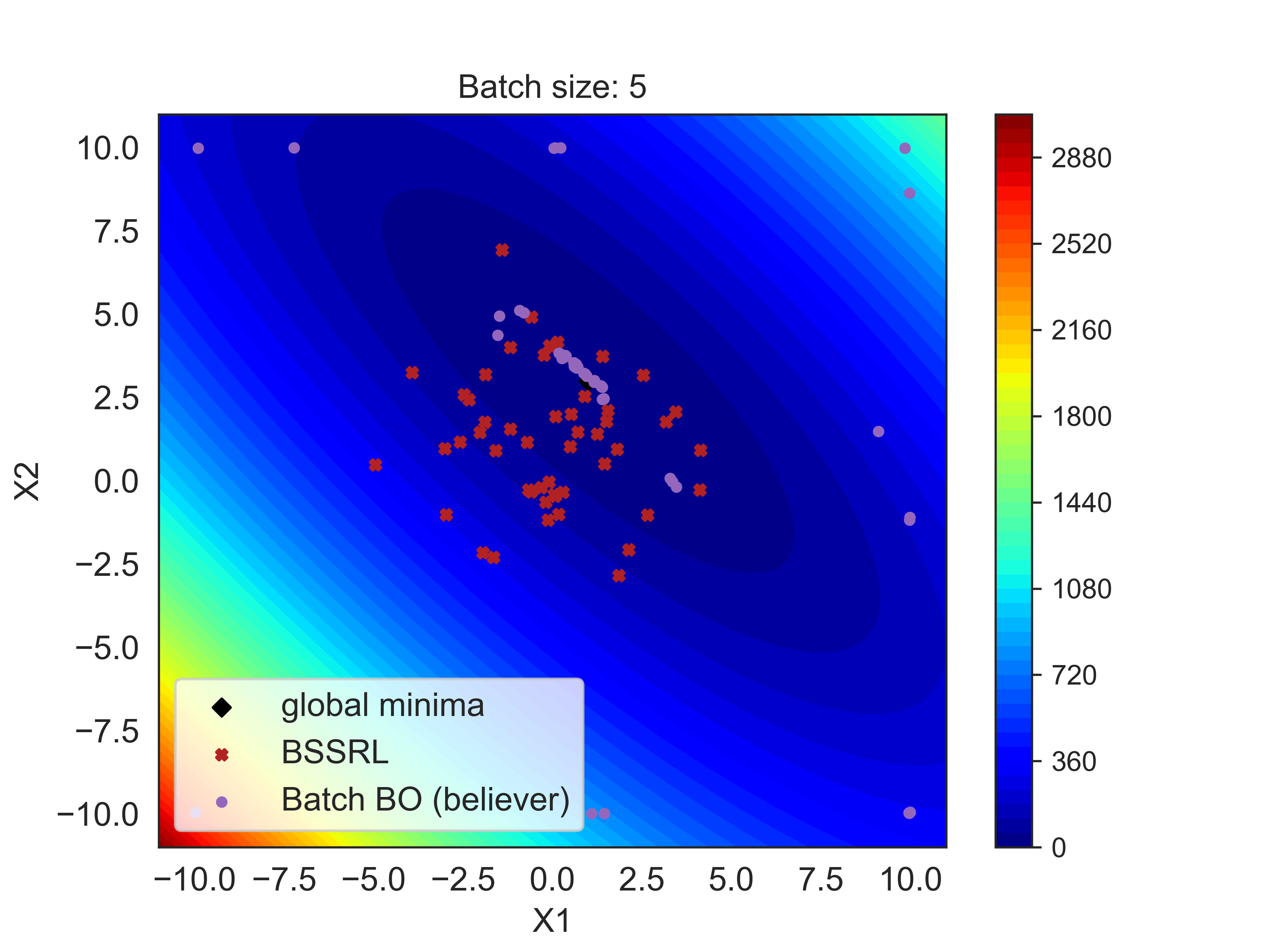}
	\includegraphics[scale=0.45]{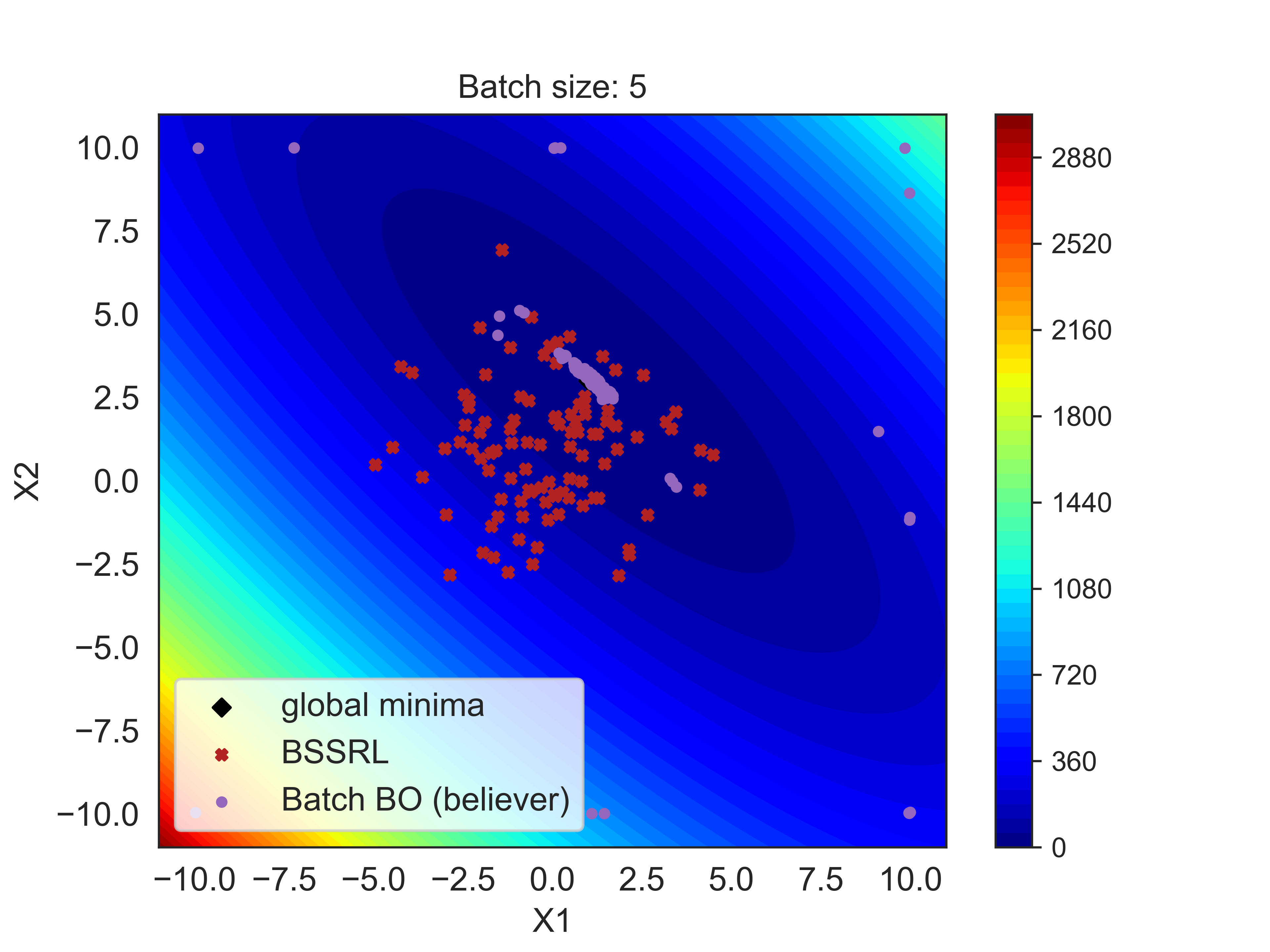}
	\caption{Designs suggested by the BSSRL and the Bayesian optimization, with batches of 5 points per batch and 20 total batches (steps).}
	\label{fig:synthetic_RL_results_3}
\end{figure}

\begin{figure}[H]
	\centering
	\includegraphics[scale=0.45]{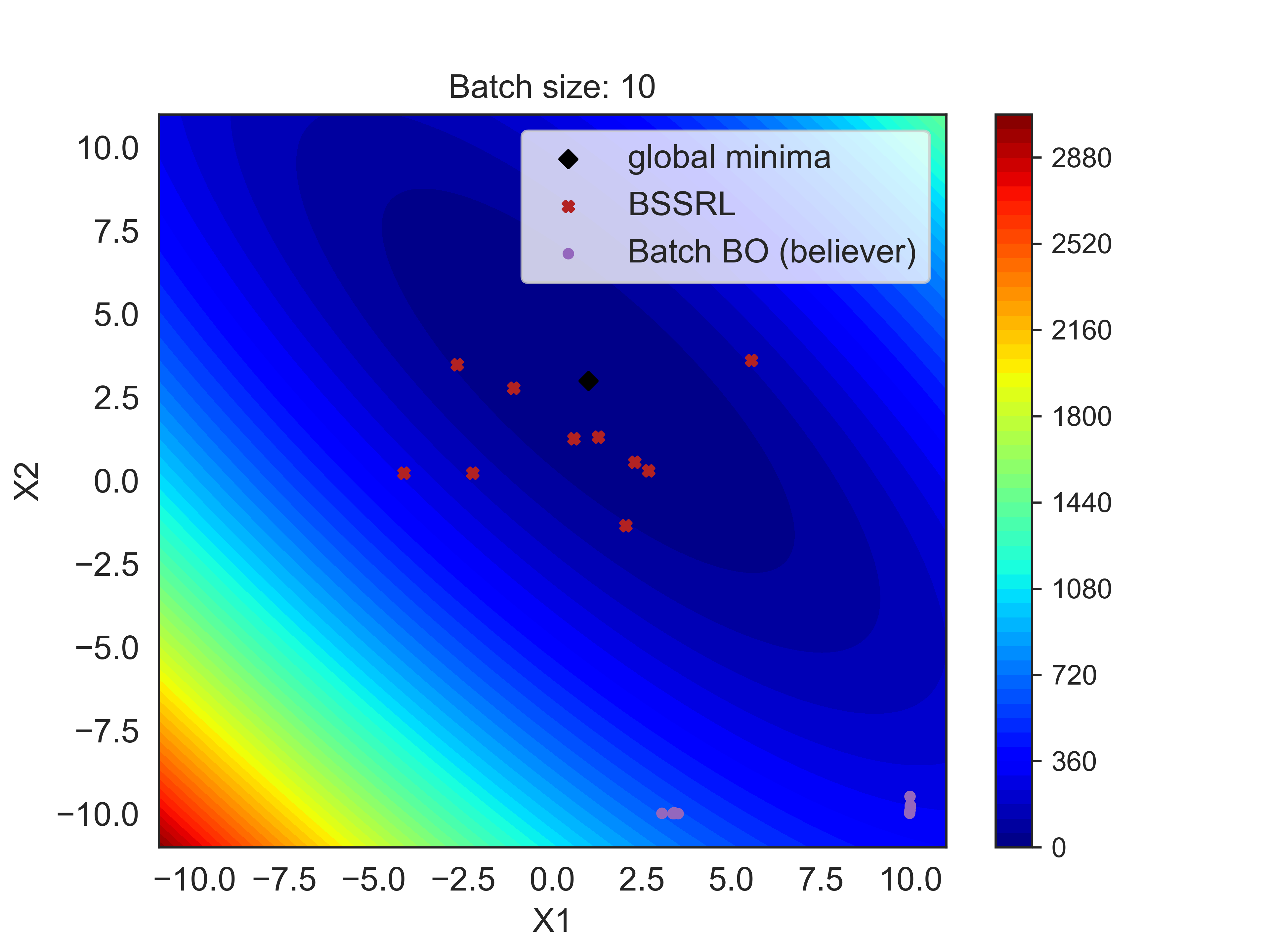}
	\includegraphics[scale=0.45]{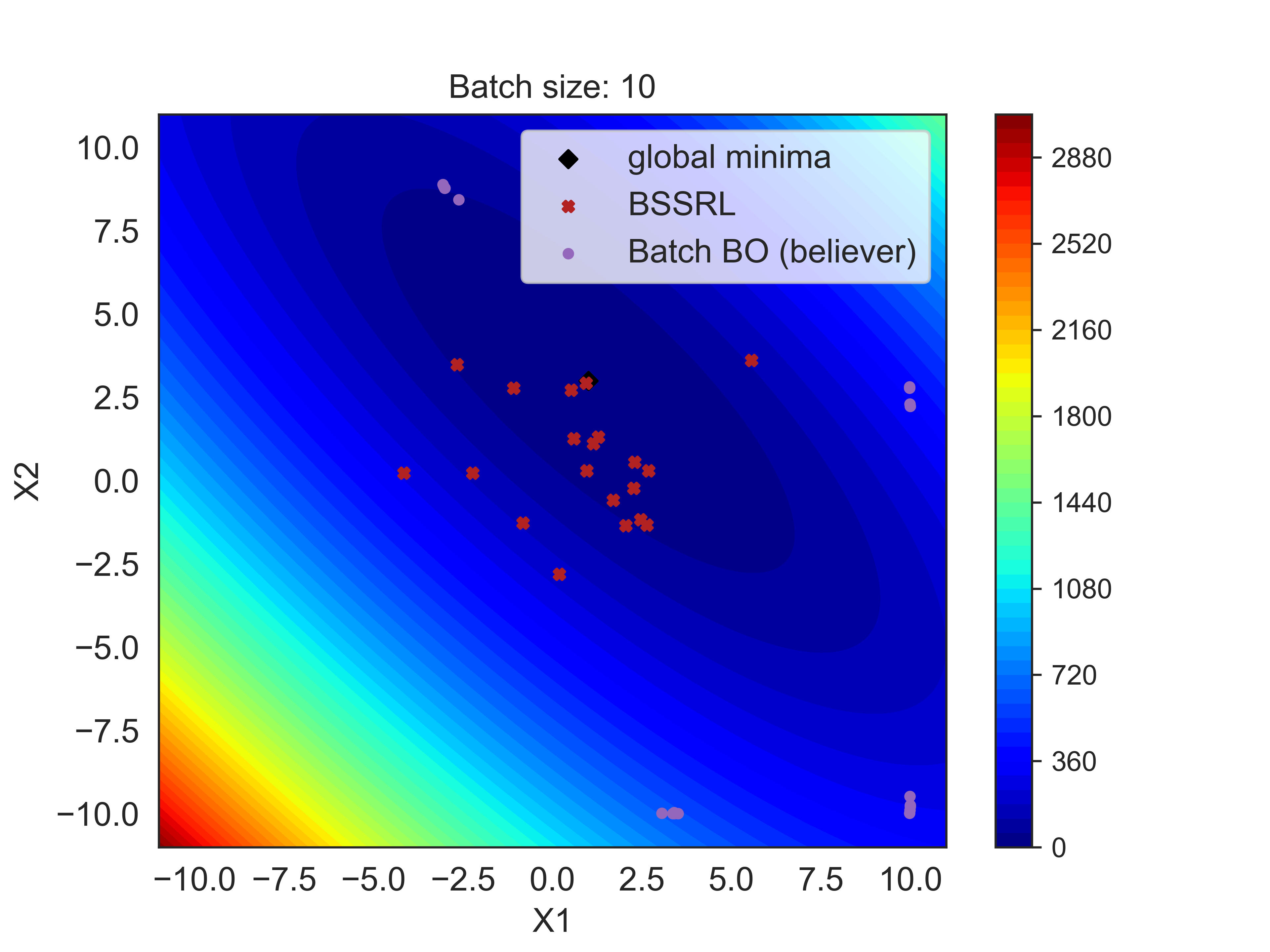}
	\includegraphics[scale=0.45]{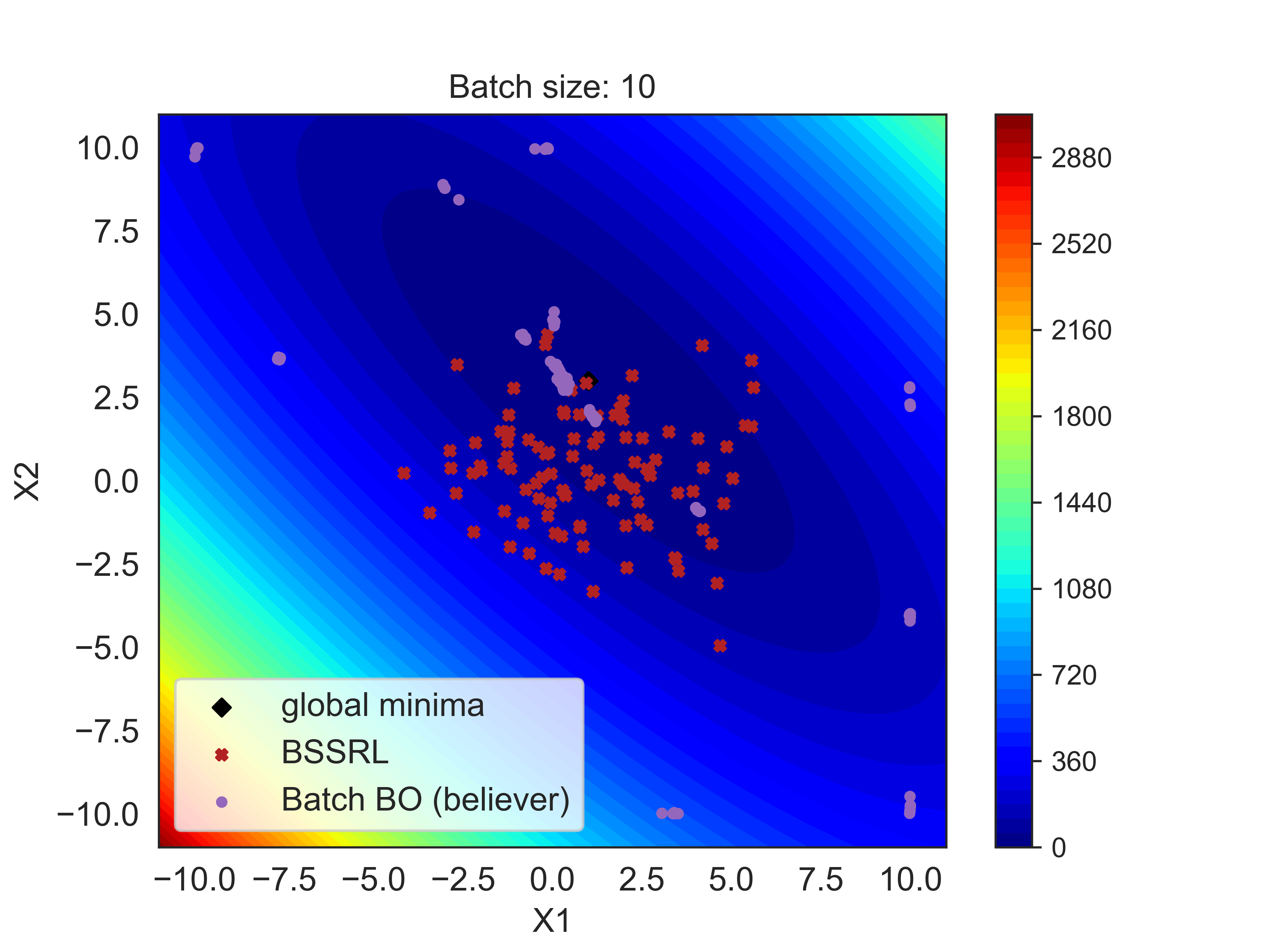}
	\includegraphics[scale=0.45]{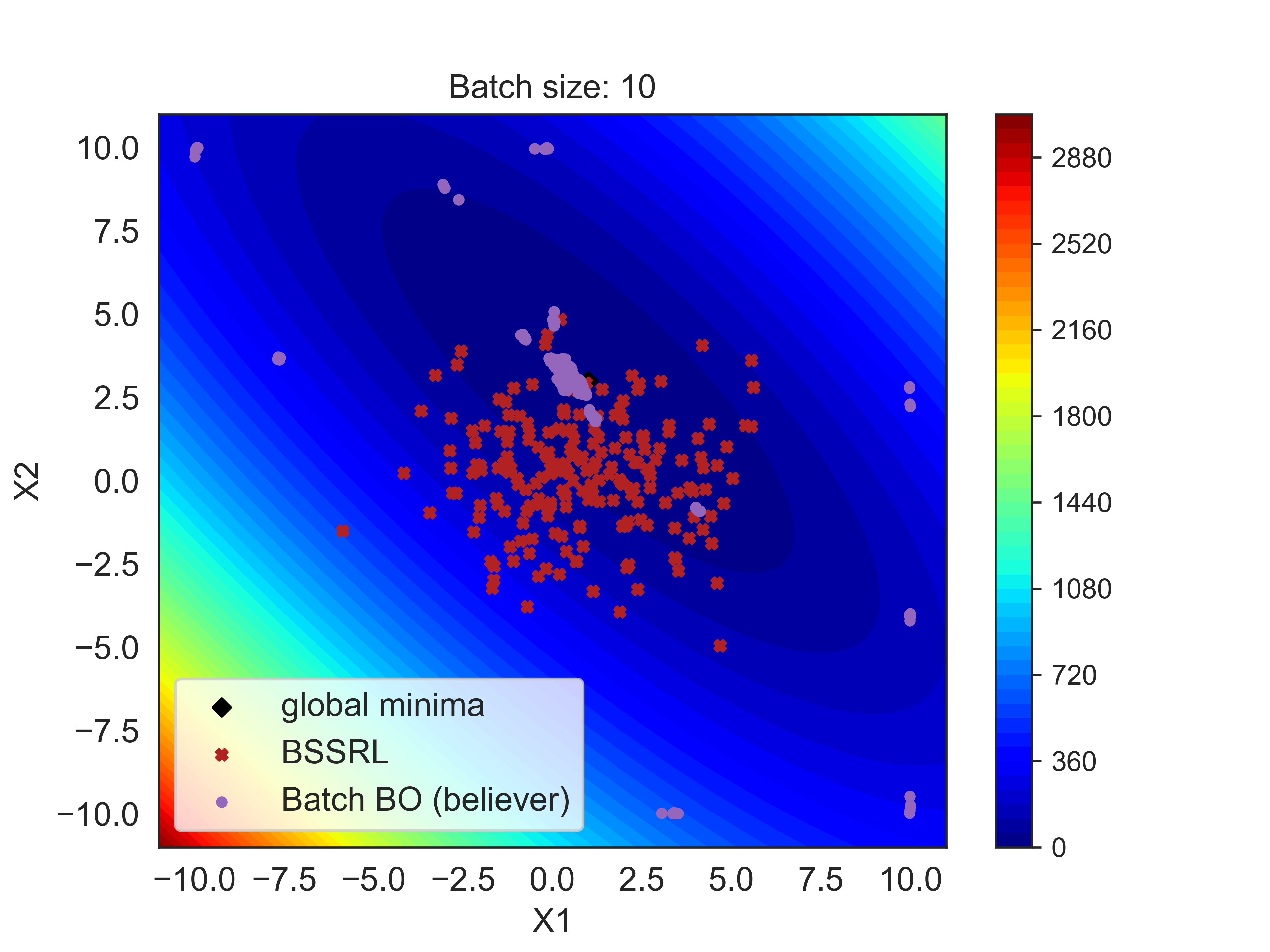}
	\caption{Designs suggested by the BSSRL and the Bayesian optimization, with batches of 10 points per batch and 20 total batches (steps).}
	\label{fig:synthetic_RL_results_4}
\end{figure}

Similar analysis for the cases where the size of the sampled batch is five and ten is shown in Figs. \ref{fig:synthetic_RL_results_3} and \ref{fig:synthetic_RL_results_4}, respectively.
In these cases, we observe a similar trend regarding the nature of the two methodologies.
The BSSRL selects batches that result in a more space-filling final set of designs, while discovering the global optimum relatively sooner.
As the size of the batch increases from three to ten, the batch BO, selects more points sub-optimal points in regions further away from the global optimum.

Even though the two methods, the BSSRL and the baseline batch BO perform relatively well and obtain multiple designs that are close to the global optimum, in some cases they do not reach the input or design corresponding to true global optimum.
This is partially due to the number of function evaluations being limited to 150, in order to enforce a practical constraint of a real engineering problem.
The other reason being the sharp gradients in the response surface of the \emph{Booth} function as a result of which most state-of-the-art optimizers would require several more samples before finding the optimum.

\subsection{3D Blade Airfoil Efficiency}
\label{sec:airfoil}

\begin{figure}[H]
	\centering
	\includegraphics[scale=0.4]{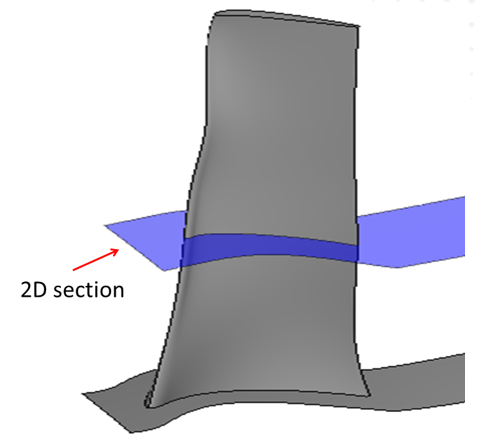}
	\caption{2D airfoil section of gas turbine last stage blade.}
	\label{fig:airfoil}
\end{figure}

In this section,  we apply the proposed algorithm for the optimization of a single 2D airfoil section of Industrial Gas Turbines (IGTs), where the performance of the airfoils are estimated using GE's in-house Computational Fluid Dynamics (CFD) tools.
The baseline design and operating conditions are derived from an industrial gas turbine last stage blade.
Figure \ref{fig:airfoil} shows the 2D section used to design the airfoil. 
The same has been extracted along a 3D streamline.
The airfoil mean radius and  the CFD domain thickness vary along the direction of the engine axis to better represent the streamtube expansion and radial mass flow distribution of the full 3D flow field. 
The airfoil shape is characterized by 12 independent parameters namely, 
the stagger angle, leading and trailing edge metal angles, leading edge diameter, suction and pressure side wedge angles, leading edge and trailing edge metal angles, and additional parameters to control the airfoil curvature between the leading and trailing edges.
The bounds on the values of these parameters have been chosen carefully, in order to allow for a deep exploration of the design space while retaining physical meaning.

The 2D airfoil flow-field is solved using TACOMA which is a GE developed solver for turbomachinery applications. 
The inlet and outlet boundary conditions are extracted from an available fully-3D stage calculation. 
Steady RANS calculations are performed to simulate the 2D airfoil flow field and characterize the turbine performance. 
A production-level grid fidelity is selected as the fine grid and is used to generate the data.
For more details on the technical aspects of the computational fluid dynamics code, we refer the interested readers to the authors' previous works~\cite{ghosh2021inverse,tsilifis2021bayesian}.

In this work, we aim to optimize the aerodynamic efficiency of the 2D airfoil. 
The aerodynamic efficiency is calculated as the ratio of mechanical and isentropic powers.
Since the data has been generated, we apply the proposed algorithm, in a retrogressive fashion, while trying to select the best design amongst the data that has been generated.
A total of 600 simulations, of input-output pairs, was used as the pre-generated data.
It is important to note that, in this case we do not do any fresh online simulations, instead we use both the methods to retrofit to the preexisting set of 600 simulations.
The minimum and maximum known values of the aerodynamic efficiency are, 94.72 and 97.39 respectively.
We begin the sequential design process with an initial dataset that contains six data points.

\begin{figure}[H]
	\centering
	\includegraphics[scale=0.45]{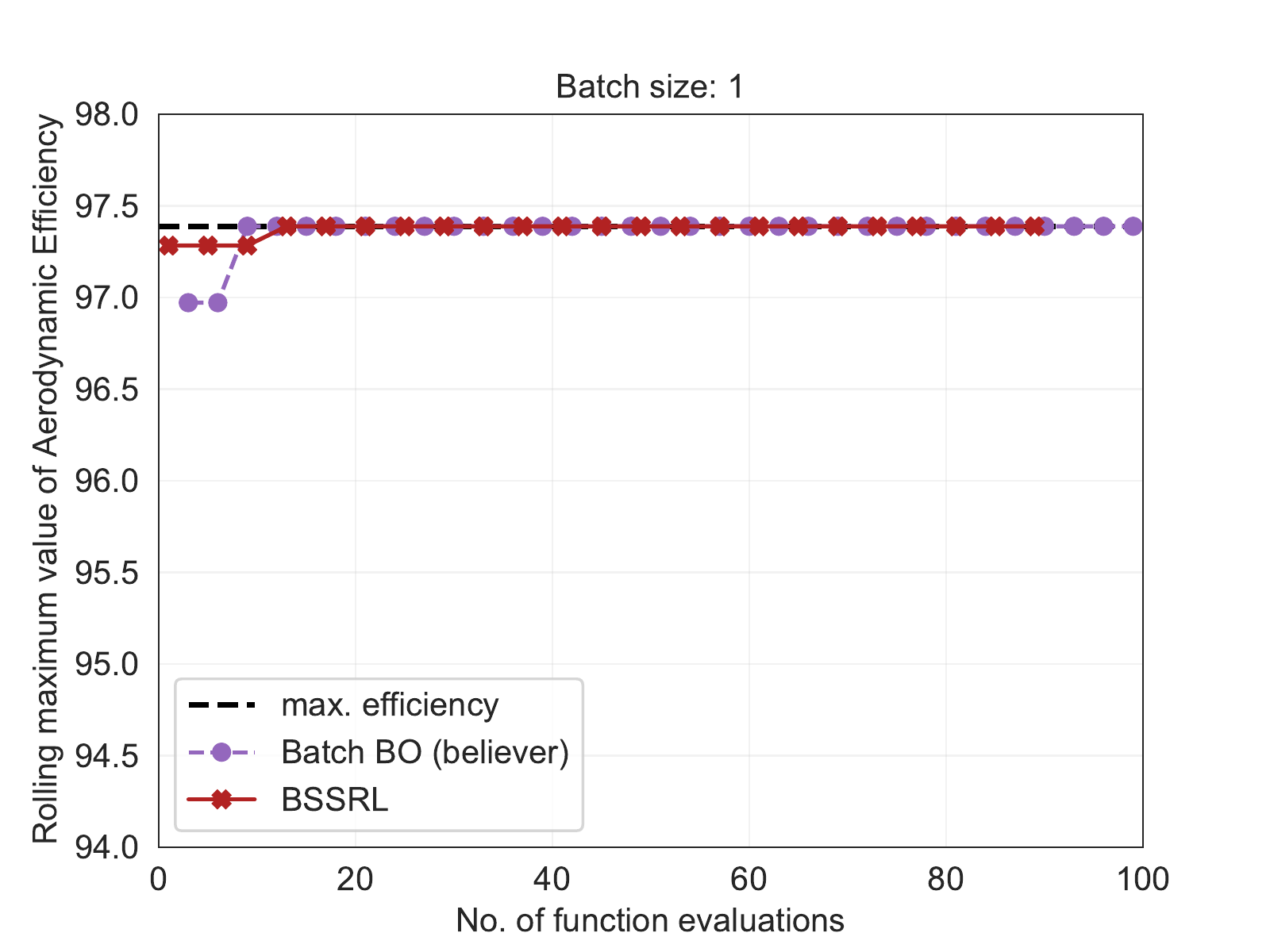}
	\includegraphics[scale=0.45]{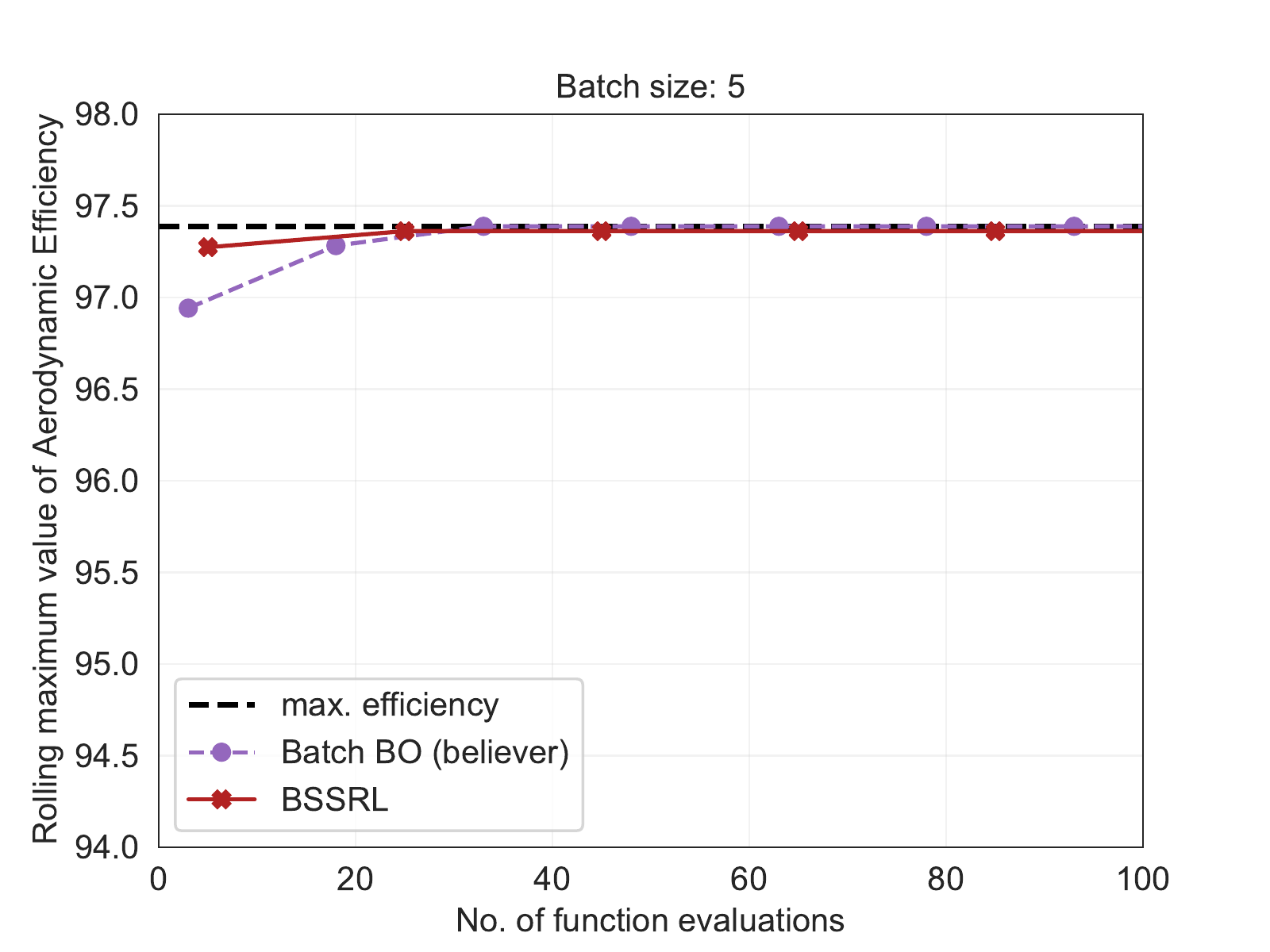}
	\includegraphics[scale=0.45]{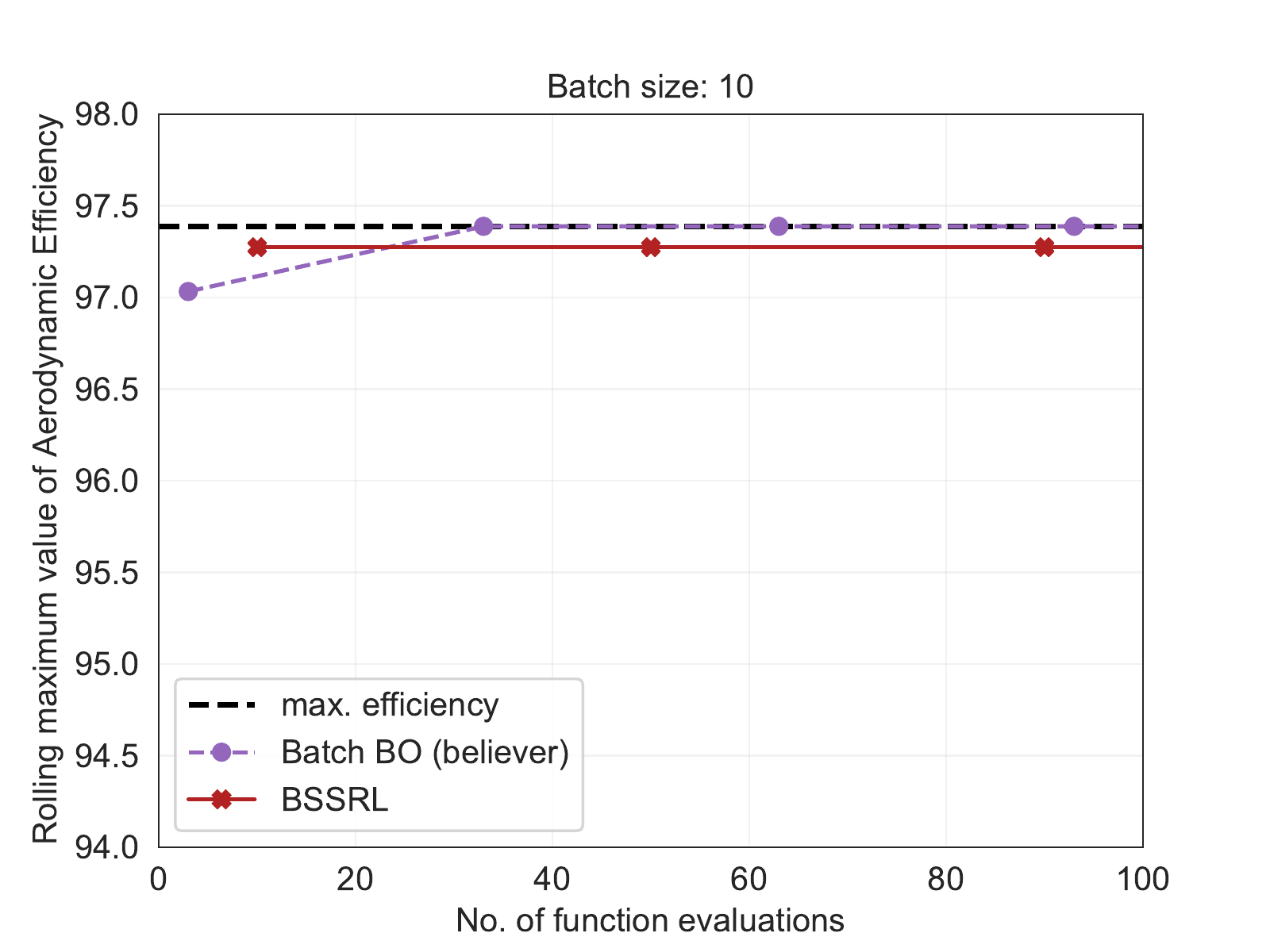}
	\caption{Convergence of the algorithm, trained on synthetic functions and applied to the 3d Airfoil design problem, with varying batch sizes. 
	Subfigure~(a) single-point per iteration, 
	~(b) 5 points per iteration, and~(c) 10 points per iteration.}
	\label{fig:airfoil_RL_results_1}
\end{figure}

Fig. \ref{fig:airfoil_RL_results_1} shows the convergence of the algorithm, subfigures show results with 200 episodes of training for the BSSRL and batches of size one, five and ten respectively. Compared to the batch BO methodology, the proposed method does reasonably well in selecting the designs close to the maximum aerodynamic efficiency.
With batches of size one and five, the BSSRL is able to select the design from the pre-generated dataset, requiring around ten points fewer than the batch BO method.

The results for this 2d airfoil problem do not really represent the actual impact of the proposed method, as the idea of doing batch sampling while selecting points from a pre-generated set of simulations is not directly compatible with the formulation of the BSSRL.
The action space, that is the batch of designs to perform the simulations, is a continous space as per the formulations presented in this paper.
While in the airfoil problem, the BSSRL can select actions that do not correspond to any of the 600 simulations.
This requires one to make certain approximations and a result of this is the delayed convergce of the BSSRL in some of the cases.
For the batch BO, this is not a problem as, it can work seamlessly on a pre-defined grid of simulations and hence the performance will be as good in this case of retro-fitting.

\section{Conclusion}
\label{sec:conclusions}
We present a framework for optimizing black-box and expensive-to-compute experiments or computer codes, where the sequence of conducting the experiments is determined by a deep RL based policy gradient method.
The uniqueness of this sequence of experiments, is the ability of sample batches of experiments, while taking into account the expected rewards for the complete budget (horizon).
The resulting algorithm, allows one to probabilistically choose the next set of actions, or experiments at each stage, while not having the need to start with a set of observed data as needed by the state-of-the-art methods in Bayesian sequential design of experiments. 
The emphasis of this model-free RL-based algorithm, is in the context of reducing the number of overall forward model evaluations, especially in scenarios where one faces logistic challenges that impede using numerous points to build an initial metamodel or from performing greedy sequential design.
The method is demonstrated on a mathematical problem of a set of functions that have similar characteristics in terms of the range of the output, dimensionality of the inputs, and computational time.
The algorithm is trained on one function, where it learns the optimal policy, that outputs a set of design points as a function of the design points from a previous iteration, while aiming to minimize the objective function.
It is then applied on the unseen test function, and the results show the effectiveness of the approach, as it learns the notorious functions' minima in less than 100 function evaluations.

The applicability of the proposed methodology, can be seen on the challenging engineering problem of 2D airfoil design of a last stage blade, where the learned policy results in reaching the maximum efficiency in a limited number of experiments when compared with a greedy myopic design of experiments scheme.
This paves the way for applying the proposed methodology to problems where one does not have the ability to train a surrogate model to start an SDOE or cannot afford enough initial experiments to train a reasonably accurate model.
The major challenges in training an optimal policy and henceforth obtaining optimal solutions for an unseen function, lie in the computational demands of the algorithm and the analytic formulation of the utility or reward function.

\section{Acknowledgment}
The authors thank Dr. Liping Wang for providing technical insights into this work and Dr. Valeria Andreoli for providing the airfoil dataset.

\bibliography{ref}
\bibliographystyle{unsrt}

\end{document}